%% file: neurips_2026.tex
\documentclass{article}

 \usepackage[main, final]{neurips_2026}

\usepackage[utf8]{inputenc} 
\usepackage[T1]{fontenc}    
\usepackage{hyperref}       
\usepackage{url}            
\usepackage{booktabs}       
\usepackage{amsfonts}       
\usepackage{nicefrac}       
\usepackage{microtype}      
\usepackage{xcolor}         
\usepackage{algorithm}
\usepackage{comment}
\usepackage{algpseudocode}
\usepackage{amsmath,amssymb}
\usepackage{multirow} 
\usepackage{xspace}
\usepackage{graphicx}    
\usepackage{caption}     
\usepackage{subcaption}  
\usepackage{float}  
\usepackage{enumerate,enumitem}
\usepackage{colortbl}
\usepackage{listings}
\usepackage{amsthm}

\theoremstyle{remark}

\definecolor{codebg}{RGB}{248,248,248}
\definecolor{codekw}{RGB}{170,13,145}
\definecolor{codecm}{RGB}{0,128,0}      
\definecolor{codedoc}{RGB}{90,90,110}   
\definecolor{codestr}{RGB}{163,21,21}
\definecolor{coderule}{RGB}{210,210,210}

\lstdefinestyle{pyseed}{
  language=Python,
  backgroundcolor=\color{codebg},
  basicstyle=\ttfamily\footnotesize,
  keywordstyle=\color{codekw}\bfseries,
  commentstyle=\color{codecm}\itshape,
  stringstyle=\color{codestr},
  morecomment=[s][\color{codedoc}\upshape\ttfamily\footnotesize]{"""}{"""},
  numbers=left, numberstyle=\tiny\color{gray},
  numbersep=8pt, frame=single, rulecolor=\color{coderule},
  framesep=6pt, breaklines=true, showstringspaces=false,
  captionpos=b, xleftmargin=14pt, xrightmargin=4pt,
  literate={->}{{$\rightarrow$}}2,
}

\setlist[itemize]{leftmargin=*}
\setlist[enumerate]{leftmargin=*}
\DeclareMathOperator*{\minimise}{minimise}

\newtheorem{defn}{\textbf{Definition}}

\newtheorem{prob}{\textbf{Problem}}

\algnewcommand\LineComment[1]{\Statex \(\triangleright\) \textit{#1}}
\newcommand{\ahd}{\texttt{AHD}\xspace}
\newcommand{\llm}{\texttt{LLM}\xspace}
\newcommand{\llms}{\texttt{LLM}s\xspace}
\newcommand{\flash}{\texttt{gemini-2.5-flash}\xspace}
\newcommand{\flashlite}{\texttt{gemini-2.5-flash-lite}\xspace}
\newcommand{\gemmasmall}{\texttt{gemma-3-12b-it}\xspace}
\newcommand{\gemma}{\texttt{gemma-3-27b-it}\xspace}

\usepackage{amsmath}
\DeclareMathOperator*{\argmax}{arg\,max}
\newcommand{\methodname}{\textsc{Ballad}\xspace}
\title{\textsc{Ballad}: Bandit-Based LLM Routing for Automated Heuristic Discovery}

\author{%
  \textbf{Samidha Verma}\textsuperscript{1} \quad
  \textbf{Ankit Anand}\textsuperscript{2} \quad
  \textbf{Sayan Ranu}\textsuperscript{1} \\[0.4em]
  {\normalfont \textsuperscript{1}Indian Institute of Technology Delhi \quad
  \textsuperscript{2}McGill University}
}

\begin{document}

\maketitle

\begin{abstract}
Modern automated heuristic discovery (\ahd) systems increasingly rival hand-crafted heuristics by leveraging Large Language Models (\llms) to iteratively mutate and refine solutions. A natural assumption is that more powerful and more expensive models consistently yield better heuristics. We challenge this empirically: across multiple NP-hard combinatorial optimization problems, we find no consistent advantage for larger, costlier models. This raises a deeper question: given a pool of \llms spanning frontier and freely hosted or open-weight models, how should one adaptively select which model to invoke at each step of the search?
The challenge is non-trivial. The optimal \llm may shift over the course of a run, and because one \llm's output seeds the context for subsequent calls, each model's reward signal is non-stationary and entangled across the heuristic population, thus violating the independence assumptions of classical online decision-making settings.  We address this through \methodname\ (\underline{\textbf{Ba}}ndit-based \underline{\textbf{LL}}M Routing for \underline{\textbf{A}}utomated Heuristic \underline{\textbf{D}}iscovery), an online bandit algorithm that learns to orchestrate a pool of \llms into an ensemble that matches or exceeds what any soloist \llm achieves alone. This orchestration is powered by an \llm selection policy that favors rare, high-scoring heuristics over stable averages, a DAG-based credit propagation mechanism through the genealogy of the heuristic population, and discounting of stale observations to remain attuned to each model's current utility. Across five NP-hard combinatorial optimization problems and a pool of four models of varying capability and cost, \methodname\ not only matches the best individual \llm, but in several cases surpasses every individual model, while reducing API cost by up to $65.6\%$
relative to always using the strongest model.
\end{abstract}

\input{intro}
\input{problem_ankit}

\input{method_ankit}
\input{experiments}

\input{results}
\input{conclusion}



\bibliographystyle{plain}
\bibliography{references}


\input{appendix}


\end{document}

%% file: intro.tex
\vspace{-0.1in}
\section{Introduction and Related Works}
\label{sec:intro}
\vspace{-0.1in}

Heuristics for hard combinatorial problems have traditionally been designed by human experts, a process that is slow, domain-specific, and difficult to scale. Recent works have shown that Large Language Models (\llms)  based systems such as FunSearch~\citep{romera2024mathematical}, EoH~\citep{liu2024evolution}, and AlphaEvolve~\citep{novikov2025alphaevolve} iteratively prompt an \llm to mutate and refine heuristics expressed as code, achieving performance that rivals or exceeds hand-crafted solutions on a range of combinatorial problems.

A practical bottleneck is cost. Automated heuristic discovery (\ahd) requires thousands of \llm calls per run, making the choice of model a first-order concern. A natural assumption is that more powerful and more expensive models consistently yield better heuristics, justifying the overhead. We challenge this assumption directly: across five combinatorial optimization benchmarks, we find \textbf{no consistent advantage for larger, costlier models}, and in several cases smaller or free models outperform their expensive counterparts. This finding has an immediate practical implication: rather than committing to a fixed model or a fixed ensemble that pays for every model on every round, one should \emph{adaptively} select which \llm to invoke at each step, based on online feedback from the search itself.

At its core, the problem maps itself to \textit{\llm routing} ~\citep{aggarwal2023automix, ong2024routellm, song2025irt, ding2025best}, where one learns to select the best LLM for a specific task and input at hand\footnote{{Please refer to App.~\ref{app:related}} for a more detailed related works discussion on \ahd and \llm routing.}. These routing approaches share two limiting assumptions: that routing decisions can be trained on offline labelled data, and that each query is routed independently with no notion of how that choice affects future calls. Neither holds in \ahd. There is no offline supervision signal; the only feedback available is the evaluator score assigned after each heuristic is tested. Also, the choice of \llm at each step in AHD affects the future routing decisions. Beyond these two limitations, AHD exposes three further properties. First, the goal is not to route each query to the best model in isolation, but to discover a single best heuristic (one that maximizes the score) across the entire run (P1). Second, because each \llm's output seeds the shared program population that conditions future prompts, routing decisions are causally coupled across steps and models (P2). Third, as the population evolves, the relative utility of each model shifts, so the optimal routing policy is non-stationary over the course of the search (P3). 

To address these challenges, we propose \methodname\ (\underline{\textbf{Ba}}ndit-based \underline{\textbf{LL}}M Routing for \underline{\textbf{A}}utomated Heuristic \underline{\textbf{D}}iscovery), which frames adaptive \llm selection in \ahd as an online bandit problem, learning which model to invoke from evaluator scores accumulated during the search itself. Classical bandit algorithms, however, assume independent arms and a cumulative reward objective, and are therefore not directly applicable under (P1) - (P3). \methodname\ addresses all three through a single per-arm index combining a bootstrapped quantile-of-maxima estimator (P1), a DAG-lineage credit signal that attributes downstream improvements to the model whose output enabled them (P2), and a sliding-window normalisation that keeps the index sensitive to each model's current rather than historical fertility (P3). \methodname\ is a drop-in replacement for the fixed-\llm generator in any FunSearch- or AlphaEvolve-style pipeline, requiring no offline supervision and no changes to prompt construction or evaluation.
\vspace{-0.05in}
\paragraph{Contributions:} \methodname makes following key contributions:
\label{sec:contributions}
\vspace{-0.02in}
\begin{enumerate}
\item \textbf{Empirical finding.} We demonstrate that larger, more expensive \llms do not consistently outperform smaller or open-weight alternatives in \ahd, motivating adaptive model routing as a practical necessity rather than a mere optimization.

\item \textbf{Problem formalization.} We formalize \llm routing in \ahd as a sequential decision problem and identify three properties: extreme-value objective (P1), cross-arm DAG coupling (P2), and non-stationary rewards (P3), that jointly distinguish it from existing bandit families.

\item \textbf{Algorithm design.} We propose \methodname, 
which casts \llm routing in \ahd as a bandit learning 
problem. \methodname is distinguished by the 
amalgamation of three algorithmic ideas tailored to the 
unique challenges of \ahd: a single per-arm index that 
(i) hunts for rare, high-scoring heuristics through 
extreme-value arm ranking rather than mean performance, 
(ii) propagates credit through a novel DAG-based model 
of the mutation genealogy, and (iii) tracks non-stationary rewards to remain attuned to each model's current rather than historical utility.\looseness=-1

\item \textbf{Empirical validation.} \methodname\ not only 
routes but orchestrates: across five NP-hard combinatorial 
optimization problems and a pool of four \llms spanning open- and closed-weight models of varying capability and cost, it matches or surpasses the best individual \llm on several problems; a result that cannot be explained by selection alone, and demonstrates that \methodname\ learns a collaborative dynamic among models that no single model can replicate. This comes at a reduced cost, cutting API expenditure by up to $65.6\%$ relative to always using the strongest model.
\end{enumerate}
\vspace{-0.12in}

%% file: problem_ankit.tex
\section{Preliminaries and Problem Formulation}
\label{sec:problem}
\vspace{-0.1in}
We begin by formalizing the automated heuristic discovery setting,  then introduce the notation needed to characterize \llm selection within it, and finally state the three properties that make this selection problem distinct from existing routing and bandit 
formulations. All notations introduced in sections ~\ref{sec:problem} and ~\ref{sec:methods} are listed in Table~\ref{tab:notations_v2} for reference.
\vspace{-0.05in}
\subsection{Automated Heuristic Discovery}
\label{sec:problem:ahd}
\vspace{-0.05in}
Given a problem $\mathcal{P}$, a set of instances 
$\mathcal{I}_{\mathcal{P}}$, and a scoring function $f_{\mathcal{P}}$, 
the goal of an \ahd system is to discover a heuristic 
$\hat{h} \in \mathcal{H}$ that maximizes average performance 
over $\mathcal{I}_{\mathcal{P}}$:
\begin{equation}
\hat{h} \;=\; \argmax_{h \in \mathcal{H}} 
\frac{1}{|\mathcal{I}_{\mathcal{P}}|}
\sum_{I \in \mathcal{I}_{\mathcal{P}}} f_{\mathcal{P}}(h, I).
\label{eq:objective}
\end{equation}
The system operates iteratively over a horizon of $T$ steps. 
At each step $t$, a context-building algorithm $\mathcal{A}$ 
selects $k$ heuristics from all heuristics generated so far, 
$H_t = \mathcal{A}(h_{<t})$, which are used to prompt an \llm 
$\ell_t \in \mathcal{L}$ to produce a new candidate $h_t$ via 
mutation and crossover. The candidate is then evaluated by a problem-specific objective function, $f_{\mathcal{P}}$, yielding a scalar score $r_t$\footnote{w.l.o.g., we assume higher is better. If lower is better, we use the negative of the score.}. We assume $\mathcal{A}$ is fixed and depends only on the set of heuristics 
available at time $t$. Traditionally, the choice 
of \llm is also fixed or decided based on a pre-decided policy. Our objective, which we will define more formally later in the section, is to \textit{learn} a policy that dynamically selects the best LLM from a given set of options at each stage of \ahd.
\vspace{-0.05in}
\subsection{Preliminary Definitions}
\label{sec:problem:definitions}
We now define the concepts that characterize the \ahd search process and that the routing algorithm will operate over.

\textbf{LLM pool.} Let $\mathcal{L} = \{\ell_1, \ldots, \ell_n\}$ 
denote the finite pool of candidate \llms. When prompted with 
context $H_t$, each \llm $\ell$ acts as a stochastic heuristic 
generator: we write $h_t \sim \mu_\ell(\ell, H_t)$ to denote a heuristic 
sampled from $\ell$'s output distribution given $H_t$. Furthermore, we denote the authoring 
\llm of $h_t$ as $\text{\llm}(h_t) \in \mathcal{L}$.

\textbf{Per-\llm score buffer.} For each \llm $\ell$, we maintain a buffer $\mathcal{O}_{\ell} = \{\!\{f_{\mathcal{P}}(h)\}\!\}$ containing the evaluation scores of all heuristics generated by $\ell$\footnote{$\mathcal{O}_{\ell}$ is a multi-set since the same heuristic may be authored multiple times by an \llm}. We denote the number of times \llm $\ell$ has 
been pulled as $n_{\ell} = |\mathcal{O}_{\ell}|$. 

\textbf{\llm quality via upper-tail quantile.} Because the 
objective in Eq.~\eqref{eq:objective} cares about the 
\emph{best} heuristic found rather than average performance, 
an \llm's value is better characterized by its ability to 
produce rare, high-scoring outputs than by its mean score. 
We therefore characterize \llm $\ell$ by its upper-tail quantile 
$q_p(\ell)$, defined as the score threshold exceeded by only 
a $p$ fraction of outputs from \llm $\ell$:
\begin{equation}
q_p(\ell) \;=\; \min\{x \in \mathbb{R} : 
\Pr(f_{\mathcal{P}}(h) > x \mid \text{\llm}(h) = \ell) \leq p\},
\label{eq:tail_quantile}
\end{equation}
where $p \in (0, 1)$ is the tail level. Smaller $p$ 
corresponds to more extreme high-score outcomes; as $p \to 0$, 
$q_p(\ell)$ approaches the maximum score achievable by \llm $\ell$, 
which is precisely the quantity our objective seeks to maximize.

\textbf{Mutation Genealogy DAG}
The key insight in AHD is that each of the LLMs don't operate independently but the output of one \llm could seed the context of other \llm, leading to coupling of different models in the system. In these situations,  a model deserves partial credit 
whenever its outputs appear as parents in the context of 
a high-scoring heuristic produced by a different model, 
even if that improvement arrives several rounds later. To operationalize this, we explicitly define the mutation genealogy of the entire heuristic population in the form of a 
a \textit{directed acyclic graph} (DAG).\\ 
\vspace{-0.1in}
\begin{defn}[Mutation Genealogy DAG]
    The genealogy of heuristics generated over the run till time $t$ is captured in the form a directed acyclic graph $\mathcal{G}_t = (\mathcal{V}, \mathcal{E})$, where each node $h \in \mathcal{V}$ is a generated heuristic and a directed edge $(h', h) \in \mathcal{E}$ indicates that $h'$ was included in the context when $h$ was generated.  
\end{defn}
\vspace{-0.05in}
For each heuristic $h$, we denote its parent set $Pa(h) \subseteq h_{<t}$ as the $k$ heuristics selected by $\mathcal{A}$ to construct the prompt from which $h$ was generated. Now, the $d$-hop ancestry of $h$ is defined recursively as the set of all heuristics reachable from $h$ by following at most $d$ parent edges:
\begin{equation}
\mathrm{Pa}^{(d)}(h) \;=\; 
\begin{cases}
\emptyset & \text{if } d = 0, \\
Pa(h) \;\cup \displaystyle\bigcup_{h' \in Pa(h)} 
\mathrm{Pa}^{(d-1)}(h') & \text{if } d \geq 1,
\end{cases}
\label{eq:ancestors}
\end{equation}
\vspace{-0.1in}
\subsection{Problem Statement}
\label{sec:problem:statement}
\vspace{-0.05in}
\begin{prob}[\textbf{Adaptive LLM Routing in AHD}]
\label{def:problem}
Given a problem $\mathcal{P}$ with instance set 
$\mathcal{I}_{\mathcal{P}}$ and scoring function 
$f_{\mathcal{P}}$, a finite pool of \llms $\mathcal{L} = 
\{\ell_1, \ldots, \ell_n\}$, a context-building algorithm 
$\mathcal{A}$, a seed heuristic $h_0$, and a finite 
horizon $T$, find an adaptive \llm selection policy 
$\pi: h_{<t} \mapsto \mathcal{L}$ that maximizes the 
score of the single best heuristic discovered across the 
entire run:
\begin{equation}
\pi^* \;=\; \argmax_{\pi} \max_{h \in h_{\leq T}^\pi} \left\{
\frac{1}{|\mathcal{I}_{\mathcal{P}}|}
\sum_{I \in \mathcal{I}_{\mathcal{P}}} 
f_{\mathcal{P}}(h, I)\right\},
\label{eq:policy_objective}
\end{equation}
where $h_{<t} = \{h_0, h_1, \ldots, h_{t-1}\}$ denotes 
the set of all heuristics generated before round $t$, 
 \llm $\ell_t = \pi(h_{<t})$ is 
selected by the policy, context $H_t = 
\mathcal{A}(h_{<t})$ is constructed by selecting $k$ 
heuristics from $h_{<t}$, and a new heuristic 
$h_t \sim \mu_{\ell_t}(\ell_t, H_t)$ is generated and 
evaluated. The seed heuristic $h_0$ is fixed and 
$\pi$ is applied for $t \geq 1$.
\end{prob}

%% file: method_ankit.tex
\vspace{-0.15in}
\section{\textsc{Ballad}: Proposed Methodology}
\label{sec:methods}
\vspace{-0.1in}
The problem of selecting the best \llm at each step $t$ can be naturally casted as an online bandit problem, where each \llm 
$\ell \in \mathcal{L}$ corresponds to an arm and the evaluator 
score $f_{\mathcal{P}}(h_t)$ serves as the reward signal. 
However, the \ahd setting departs from standard bandit 
formulations in three fundamental ways, each of which 
invalidates a core assumption of classical bandit algorithms 
and necessitates a tailored solution.
\vspace{-0.1in}
\begin{enumerate}
\item \textbf{[P1] Extreme-value objective.} The 
goal is to maximize the score of the single best heuristic 
found across all $T$ steps, not the expected score of any 
individual \llm call. Arms should therefore be ranked by 
their upper-tail quality $q_p(\ell)$ rather than their 
mean, and $q_p(\ell)$ must be estimable reliably from the 
small buffer $\mathcal{O}_{\ell}$ accumulated during the run.

\item\textbf{[P2] Cross-arm coupling.}  \ahd relies fundamentally on mutation: at each step, an \llm is prompted with a set of existing heuristics and asked to produce an improved variant. When a single \llm drives the entire search, attributing reward to that \llm is unambiguous. With a pool of \llms, however, this attribution becomes non-trivial. When a strong heuristic is discovered, should the reward \textit{only} go to the \llm that produced it, or also to those whose outputs contributed to its prompt? The programs in the prompt may have been authored by different \llms in earlier rounds, making the discovery a combinatorial outcome of multiple models' contributions across multiple steps. Attributing reward solely to the calling \llm is therefore blind to this combinatorial nature of the discovery process: it ignores the possibility that a weaker \llm produced a heuristic that, when placed in the context of a stronger one, unlocked a breakthrough that neither model could have achieved alone. Any arm quality estimator that ignores this coupling will  misevaluate the true contribution of 
each model to the search.

\item\textbf{[P3] Non-stationary reward 
distributions.} The score distribution $P_t(\cdot \mid \ell)$ 
of \llm $\ell$ drifts with $t$ as the population of heuristics 
evolves. An estimator that pools all past observations of 
\llm $\ell$ uniformly will lag behind the current regime, 
over-weighting contributions from early rounds when the 
population was less refined.

\end{enumerate}

No existing \llm router or bandit algorithm simultaneously 
satisfies all three requirements and hence, we design \methodname.
\vspace{-0.1in}
\subsection{Design Overview}
\label{sec:method:overview}
\methodname addresses the three properties (P1) - (P3) 
 through a single per-arm scalar score $\mathrm{Score}_{\ell}(t)$ that is recomputed at every round from evaluator scores accumulated till the current timestep of the search process. Each \llm $\ell\in\mathcal{L}$ corresponds to an arm and at any round $t$, the arm with the highest \textit{score} is pulled, i.e., $\pi^*=\arg\max_{\ell\in\mathcal{L}}Score_{\ell}(t)$. The score has three 
additive terms, each targeting a specific characteristic, to holistically characterize the current state of the search process:
\begin{equation}
\boxed{\;
\mathrm{Score}_{\ell}(t)
\;=\;
\underbrace{(1-\lambda)\,\hat{q}_\tau(\ell) + 
\lambda\,\max\{\mathcal{O}_{\ell}\}}_{\text{hybrid extreme-value (P1)}}
\;+\;
\underbrace{\alpha\,\bar{L}_{\ell}(t)}_{\text{Mutation genealogy 
+ recency (P2,\,P3)}}
\;+\;
\underbrace{\beta\,\sqrt{\frac{\log T} { n_{\ell}}}}_{\text{exploration}}
\;}
\label{eq:score}
\end{equation}
The first term ranks arms by their ability to produce rare, 
high-scoring heuristics rather than by average performance, 
where $\hat{q}_\tau(\ell)$ is a bootstrap estimate of the 
upper-tail quantile $q_p(\ell)$ of \S\ref{sec:problem:definitions} 
computed from the finite sample $\mathcal{O}_{\ell}$, with 
$\lambda \in [0,1]$ controlling the blend between this 
robust tail estimator and the empirically recorded maximum 
$\max \{\mathcal{O}_{\ell}\}$.
The second term rewards \llms whose past outputs have seeded 
improvements by other \llms in later rounds, capturing 
the combinatorial nature of multi-\llm discovery, 
expressed as a windowed fertility score $\bar{L}_\ell(t)$ 
(defined in \S\ref{sec:method:lineage}), with $\alpha 
\geq 0$ controlling its contribution relative to the base 
score. The third is an exploration bonus weighted by $\beta \geq 0$ that prevents the score from prematurely concentrating pulls on a single arm, ensuring that every model accumulates enough observations for its score to be reliable before exploitation dominates. 
We next deep-dive into each of the individual terms in
\S\ref{sec:method:qomax}--\S\ref{sec:method:ucb}.
\vspace{-0.1in}
\subsection{Hybrid Extreme-Value Base Score (P1)}
\label{sec:method:qomax}
\vspace{-0.1in}
The first term replaces the empirical mean of $\mathcal{O}_{\ell}$ 
with an upper-tail estimator that is robust at small sample 
sizes. Two failure modes motivate the design. A direct plug-in 
estimator of a high quantile of $\mathcal{O}_{\ell}$ has high 
variance when $n_{\ell}$ is small, making it unreliable early in 
the run. Conversely, $\max \mathcal{O}_{\ell}$ is a single order 
statistic that can be dominated by one outlier draw, 
overstating an arm's true tail quality. Neither alone is 
sufficient; we therefore combine them through a bootstrap.

Given the score buffer $\mathcal{O}_{\ell}$ of size $n_{\ell}$, we 
draw $B$ bootstrap resamples of scores in
$\mathcal{O}_{\ell}^{(b)} \sim \mathrm{Unif}(\mathcal{O}_{\ell}, n_{\ell})$, take the per-resample maximum 
$M_{\ell}^{(b)} = \max \mathcal{O}_{\ell}^{(b)}$, and estimate
\begin{equation}
\hat{q}_\tau(\ell)
\;=\;
\mathrm{Quantile}_\tau\!\bigl(\{M_{\ell}^{(b)}\}_{b=1}^{B}\bigr),
\label{eq:qalpha}
\end{equation}
where $\tau \in (0,1)$ is the bootstrap quantile level. 
$\hat{q}_\tau(\ell)$ estimates the upper-tail quantile 
$q_p(\ell)$ of \S\ref{sec:problem:definitions} but with 
greater stability: it is the $\tau$-quantile of the 
bootstrap distribution of the sample maximum, which remains 
well-behaved under heavy tails even at small $n_{\ell}$ 
\citep{baudry2022efficient}. Yet discarding 
$\max \mathcal{O}_{\ell}$ entirely would discard the single most 
informative observation, since the best heuristic found is 
exactly what \ahd reports at the end of the run. We therefore 
form the hybrid estimator:
\begin{equation}
\mathrm{base}(\ell)
\;=\;
(1-\lambda)\,\hat{q}_\tau(\ell)
\;+\;
\lambda \max \mathcal{O}_{\ell},
\qquad \lambda \in [0,1],
\label{eq:base}
\end{equation}
which interpolates between a low-variance tail estimator 
($\lambda = 0$) and a record-value estimator ($\lambda = 1$). 
The sensitivity of performance to $\lambda$ is examined in 
the ablation study (\S~\ref{sec:ablations}).
\vspace{-0.05in}
\subsection{Lineage Credit on the Heuristic DAG (P2)}
\label{sec:method:lineage}
\vspace{-0.1in}
The second term focuses on rewarding cross-arm coupling. Using heuristic genealogy graph $\mathcal{G}_t$ defined in \S~\ref{sec:problem:definitions}, whenever a  heuristic $h$ is discovered, we traverse its ancestry in 
$\mathcal{G}_t$ and attribute a credit $c$ to every \llm that occurs in the ancestors. This transforms the routing problem from one of evaluating isolated \llm outputs to one of reasoning about the causal influence of each model across the entire search trajectory. However, since there are only a few LLMs which generate thousand of heuristics, a na\"ive credit assignment in time to all ancestors could lead to almost uniform credit assignment across all \llms. To address this, we introduce restrictions across three different axes in credit assignment, across depth, time and absolute score respectively. Each of these restrictions help in introducing novel differentiation in \llm arms leading to significant performance improvements.  Mathematically,  let $h_t$ denote the heuristic generated at  round $t$, and let $\mathrm{Pa}^{(d)}(h_{t})$ denote 
the set of ancestor heuristics within $d$-hop from $h_t$ in 
$\mathcal{G}_t$. 
For each arm $\ell \in \mathcal{L}$, we maintain a (timestamped) credit history $\mathcal{C}_\ell$ as an ordered sequence of $(t', c)$ pairs, where $t' \leq t$ is the round at which credit $c$ was earned due to lineage. Credits are stored with their birth timestamp $t$, which enables accounting for recency. 

\vspace{-0.1in}
\paragraph{Adaptive Clipped Reward:} Firstly, we focus on absolute score.  The reward of a newly generated heuristic $h_t$ denoted as $r_t$ could be set to simply $f_{\mathcal{P}}(h_t)$. However, this risks flooding the lineage 
signal with routine improvements from, wherein in reality, we are interested in improvements that pushes the frontier. To capture this, we clip each new score against a dynamic baseline. Specifically, the reward $r_t$ is computed as $r_t \;=\; \max\bigl(0,\; f_{\mathcal{P}}(h_t) - b_t\bigr)$ where $b_t$ is a time dependent dynamic baseline. Specifically, we use median elite baseline, $b_t=\mathrm{median}(\mathrm{top}_K\!\bigl(\bigcup_{\ell \in 
\mathcal{L}} \mathcal{O}_\ell\bigr))$ where $\mathrm{top}_K$ denote the top-K scores in the set.

%
Credit is propagated only when $r_t > 0$, i.e.\ when 
$h_t$ genuinely advances beyond the current elite 
frontier. The median of the elite pool is a natural threshold: it rises adaptively as the search progresses, ensuring that 
only heuristics that genuinely advance the frontier 
propagate credit, while keeping the threshold robust to 
outliers. Setting $b_t$ too low would flood ancestor arms 
with uninformative credit; setting it too high would starve the lineage signal. Using the elite median balances these two failure modes. This design is consistent with the elitist parent selection used in FunSearch~\citep{romera2024mathematical}, AlphaEvolve~\citep{novikov2025alphaevolve}, and ReEvo~\citep{ye2024reevo}.
 
\vspace{-0.1in}
\paragraph{Decaying credit with genealogical distance}
As the genealogy graph $\mathcal{G}_t$ grows to hundreds of nodes, the causal influence between different nodes in the same lineage decreases with each mutation. This is captured in credit assignment by assigning a credit $c = \gamma^d \cdot r_t $ equivalent to a geometrically-discounted share of the reward $r_{t}$ obtained from $h_t$ to every ancestor arm, with discount factor $\gamma \in (0,1)$.  Simply, this ensures that direct parents receive more credit than distant ancestors. In addition, credits are bounded up to depth $D_{\max}$ from the newly generated heuristic.
\begin{equation}
\mathcal{C}_{\mathrm{\llm}(h')} \;\leftarrow\; 
\mathcal{C}_{\mathrm{\llm}(h')} \;\cup\; 
\bigl\{\bigl(t,\; \gamma^d \cdot r_t\bigr)\bigr\}
\qquad
\forall h' \in \mathrm{Pa}^{d}(h_t); 0 \leq d \leq D_{\max}
\label{eq:bfs}
\end{equation}

\vspace{-0.1in}
\paragraph{Recency bias (P3)}
\label{sec:method:window}
With the credit attribution rule of 
Eq.~\eqref{eq:bfs} in place, a natural choice for 
the lineage score of arm $\ell$ at round $t$ is simply 
the sum of all credits accumulated in 
$\mathcal{C}_l$, i.e.,
 $L_\ell(t) \;=\; \sum_{(t', c') \in \mathcal{C}_\ell} c'$. This is intuitive but problematic: it accumulates credit 
indefinitely, so an arm that was highly productive early 
in the search, when the heuristic population was crude 
and easy to improve upon, would maintain a high  score long after the population had moved on and its 
outputs were no longer competitive. To address this, we restrict credit assignment in time axis. Effectively, the Lineage credit $L_l(t)$ is restricted to credits earned within the last $W$ rounds, i.e., $\mathcal{C}^W_\ell \;\leftarrow\;
\bigl\{(t', c') \in \mathcal{C}_\ell : t' > t -W\bigr\}$,
and define the windowed fertility score as:
\begin{equation}
\bar{L}_\ell(t)
\;=\;
\frac{\displaystyle\sum_{(t',c') \in \mathcal{C}^W_\ell} 
c'}{\log\!\bigl(\max(e,\, n_\ell)\bigr)}.
\label{eq:fertility}
\end{equation}
The logarithmic denominator in 
Eq.~\eqref{eq:fertility} corrects for exposure 
bias: an arm pulled more frequently has contributed 
more programs to the population and therefore had more 
opportunities to appear in parent sets; not necessarily 
because it is a better model, but simply because it has 
been used more. Dividing by $\log(n_\ell)$ corrects for 
this. The floor at $\log e = 1$, where $e$ denotes the Euler's number, prevents arms with very few pulls from having their lineage score artificially inflated by division by a near-zero quantity. 
\vspace{-0.1in}
\subsection{Exploration Bonus and Selection Rule}
\label{sec:method:ucb}
\vspace{-0.1in}
The base score and lineage term are both point estimates 
derived from observed data. In the absence of 
continued testing of under-explored arms, the score would 
concentrate pulls on whichever \llm produced a high early 
score, leaving other \llms starved and potentially 
missing a stronger one whose quality only becomes apparent after sufficient sampling. To overcome this, we add a fixed-horizon UCB bonus $\beta \sqrt{\frac{\log T}{n_\ell}}$ as the third term to Eq.~\ref{eq:score}.

We use the total horizon $T$ rather than the current 
round $t$ in the numerator. This is deliberate: because 
the budget is fixed and known in advance, the exploration 
bonus is calibrated to the full horizon rather than growing 
unboundedly with elapsed time. Concretely, this means the 
bonus shrinks faster than standard UCB by front-loading 
exploration when the budget is plentiful and concentrating 
exploitation in later rounds when the remaining budget is 
scarce. 

\vspace{-0.1in}
\subsection{Pipeline integration.}
\vspace{-0.1in}
\methodname is a drop-in replacement for the fixed-\llm 
generator inside any FunSearch- or AlphaEvolve-style 
loop~\citep{romera2024mathematical, novikov2025alphaevolve}. 
The host framework owns prompt construction, evaluation, 
and the program database; \methodname is touched at exactly 
two points per round. Before prompt construction, the 
router selects which \llm to invoke based on 
$\mathrm{Score}_\ell(t)$ as described above. After 
evaluation, the router updates the selected arm's score 
buffer $\mathcal{O}_\ell$ with $f_{\mathcal{P}}(h_t)$, 
increments $n_\ell$, and propagates lineage credits 
through $\mathcal{G}_t$ if $r_t > 0$. No information 
beyond the evaluator score $f_{\mathcal{P}}(h_t)$, the ancestry set $Pa^{Dmax}(h_t)$ and 
the identity of the selected \llm needs to cross the 
interface, making \methodname lightweight and agnostic 
to the specifics of prompt construction and evaluation. Pseudocodes and time complexity analysis are provided in App.~\ref{app: details}.

%% file: experiments.tex
\vspace{-0.1in}
\section{Experiments}
\label{sec:exps}
\vspace{-0.1in}
In this section, we benchmark \methodname on NP-hard 
combinatorial optimization (CO) problems to answer the 
following concrete research questions:
\vspace{-0.005in}
\begin{itemize}
    \item \textbf{RQ1. Does model size predict 
    performance?} Is the common assumption that ``larger, 
    more expensive \llms consistently yield better 
    heuristics'' empirically supported?
    \item \textbf{RQ2. Matching vs.\ surpassing.} Does 
    \methodname learn to orchestrate the pool of \llms 
    such that the resulting ensemble not only matches 
    but \textit{exceeds} the best individual \llm in 
    hindsight?
    \item \textbf{RQ3. Cost effectiveness.} Does 
    \methodname achieve competitive or superior 
    performance at a lower API cost than always using 
    the strongest model?
    \item \textbf{RQ4. Ablation.} What are the marginal 
    contributions of each component of \methodname's 
    index score (Eq.~\ref{eq:score}), and how sensitive is performance to 
    the choice of hyperparameters?
\end{itemize}
The codebase of \methodname and the programs generated for the various NP-hard CO problems are available at \url{https://github.com/idea-iitd/Ballad}. Hardware details and hyperparameter settings are described in App.~\ref{app:reproducibility}.

\vspace{-0.05in}
\subsection{Experiment Setup}
\label{sec: setup} 
\vspace{-0.01in}
\paragraph{Problems.} 
We evaluate \methodname across $3$ classes of NP-hard combinatorial optimisation problems.\looseness=-1
\vspace{-0.01in}
\begin{itemize}
    \item \textbf{Routing} includes the Traveling 
Salesman Problem (TSP) and Capacitated Vehicle Routing 
Problem (CVRP), both on instances with 50 nodes. CVRP uses a vehicle capacity constraint of 
$W=50$.
\item \textbf{Bin packing} includes both the Online 
and Offline Bin Packing Problems. The online variant 
uses large-scale instances of 5,000 items drawn from a 
Weibull distribution~\citep{liu2024evolution, 
zhengmonte, castineiras2012weibull}, evaluated under 
two bin capacity settings ($C \in \{100, 500\}$); the 
offline variant uses 500-item instances with bin 
capacity $C=150$.
\item \textbf{Graph Edit Distance (GED)} 
is evaluated on the Linux dataset~\citep{linux_dataset}, 
comprising unlabelled program dependence graphs with an 
average of 8 nodes and 7 edges per graph. 
\end{itemize}  
\vspace{-0.1in}
Detailed problem descriptions are provided in 
App.~\ref{app:problem_description}; dataset splits and 
instance counts are summarized in 
Table~\ref{tab:dataset_comprehensive}.

\textbf{LLMs.} We consider $4$ models for this study, \flashlite~\cite{comanici2025gemini} and \flash~\cite{comanici2025gemini} as the two closed-source paid models, and \gemmasmall~\cite{Kamath2025Gemma3T} and \gemma~\cite{Kamath2025Gemma3T} as the two open-weight free models. Among these, \flash is the largest and most expensive. 

\textbf{Metrics:} We benchmark \methodname along three dimensions: \textbf{(i)} the average \textbf{objective value} attained by each \llm and \methodname, \textbf{(ii)} the average \textbf{rank} across runs for each \llm and \methodname, and \textbf{(iii)} the inference (API) \textbf{cost} incurred by each \llm and \methodname. All experiments are repeated three times; we report the mean and standard deviation. Both the objective value and the rank are evaluated on the held-out test set. For each method, the selected heuristic is the one that achieves the best objective value on the validation set. The API cost for \ahd is measured on the training set. Details of the train/validation/test splits are provided in Table~\ref{tab:dataset_comprehensive}. 

%% file: results.tex
\vspace{-0.1in}
\subsection{RQ1: Does model size predict 
    performance?}
\label{sec:rq1}
\vspace{-0.1in}
Table~\ref{tab:test_comparison} reports the objective values and ranks achieved by each \llm and \methodname. We observe that \flash, despite being the largest and most capable model, does not achieve the best objective value or rank on any of the problems. In contrast, \gemmasmall, the smallest model, achieves the best performance on CVRP. Furthermore, \flashlite exhibits the most consistent performance across all combinatorial optimization tasks.

These results clearly indicate that larger models do not necessarily yield better performance, thereby motivating the need for \llm routing in \ahd.

\begin{table}[t]
\vspace{-0.3in}
\centering
\small
\caption{Test-set performance. We report the best program (selected on validation) evaluated on the test set. For each model, we show mean $\pm$ std (over 3 runs) for \textit{Task-Specific Objective (Obj.)} ($\downarrow$) and \textit{Rank} ($\downarrow$), where lower is better for both Obj, since we benchmark on minimization problems, and rank. Ranks are integer-valued; draws at full precision share the lowest rank, and the next rank is skipped. \textit{Overall} is the mean of per-problem means; its std is the root-mean-square of per-problem stds.}
\label{tab:test_comparison}
\scalebox{0.75}{%
\begin{tabular}{llcccccc}
\toprule
\textbf{Model} & \textbf{Metric} & \textbf{TSP} & \textbf{CVRP} & \textbf{Online BP} & \textbf{Offline BP} & \textbf{GED} & \textbf{Overall} \\
\midrule
gemma-3-12b-it & Obj. & 6.396 $\pm$ 0.028 & \underline{10.753 $\pm$ 0.031} & 2.510 $\pm$ 0.000 & 215.719 $\pm$ 0.000 & 4.953 $\pm$ 0.091 & 48.066 $\pm$ 0.045 \\
 & Rank & 3.333 $\pm$ 0.471 & \textbf{1.667 $\pm$ 0.471} & 3.000 $\pm$ 0.000 & 4.000 $\pm$ 0.000 & 4.667 $\pm$ 0.471 & 3.333 $\pm$ 0.365 \\
\midrule
gemma-3-27b-it & Obj. & 6.934 $\pm$ 0.003 & 15.735 $\pm$ 4.348 & 2.593 $\pm$ 0.059 & 215.719 $\pm$ 0.000 & 4.804 $\pm$ 0.040 & 49.157 $\pm$ 1.945 \\
 & Rank & 4.667 $\pm$ 0.471 & 5.000 $\pm$ 0.000 & 4.333 $\pm$ 0.943 & 4.000 $\pm$ 0.000 & 2.333 $\pm$ 0.471 & 4.067 $\pm$ 0.516 \\
\midrule
gemini-2.5-flash-lite & Obj. & \underline{6.258 $\pm$ 0.064} & 10.776 $\pm$ 0.070 & \underline{1.574 $\pm$ 0.217} & \underline{204.750 $\pm$ 0.034} & \underline{4.792 $\pm$ 0.032} & \underline{45.630 $\pm$ 0.108} \\
 & Rank & \textbf{1.333 $\pm$ 0.471} & 2.333 $\pm$ 0.943 & \underline{1.667 $\pm$ 0.471} & \textbf{1.667 $\pm$ 0.471} & \textbf{1.667 $\pm$ 0.943} & \textbf{1.733 $\pm$ 0.699} \\
\midrule
gemini-2.5-flash & Obj. & 6.720 $\pm$ 0.304 & 11.984 $\pm$ 0.743 & 2.510 $\pm$ 0.000 & 204.781 $\pm$ 0.058 & 4.941 $\pm$ 0.069 & 46.187 $\pm$ 0.361 \\
 & Rank & 3.667 $\pm$ 1.247 & 4.000 $\pm$ 0.000 & 3.000 $\pm$ 0.000 & 2.333 $\pm$ 0.943 & 4.333 $\pm$ 0.471 & 3.467 $\pm$ 0.730 \\
\midrule
\methodname & Obj. & \textbf{6.233 $\pm$ 0.056} & \textbf{10.685 $\pm$ 0.075} & \textbf{1.483 $\pm$ 0.276} & \textbf{204.693 $\pm$ 0.323} & \textbf{4.789 $\pm$ 0.022} & \textbf{45.577 $\pm$ 0.195} \\
 & Rank & \underline{1.667 $\pm$ 0.471} & \underline{2.000 $\pm$ 0.816} & \textbf{1.333 $\pm$ 0.471} & \underline{1.667 $\pm$ 0.943} & \underline{2.000 $\pm$ 0.816} & \underline{1.733 $\pm$ 0.730} \\
\bottomrule
\end{tabular}%
}
\vspace{-0.2in}
\end{table}

\vspace{-0.1in}
\subsection{RQ2: Can \methodname match or surpass the best \llm?}
\label{sec:rq2}
\vspace{-0.05in}
Revisiting Table~\ref{tab:test_comparison}, we find that \methodname achieves the best objective value overall and is consistently ranked first or second across all tasks. Aggregated across problems, it matches or outperforms all individual \llms in both average objective and rank.

These results highlight two key observations. First, the strong aggregate performance of \methodname indicates robust behavior across heterogeneous problem instances, even when the identity of the best-performing \llm varies by task. This suggests that \methodname effectively learns to select an appropriate sequence of \llm calls conditioned on the problem. Second, and more notably, \methodname occasionally surpasses the best single \llm on individual tasks. This indicates that it is not merely selecting the best model, but composing multiple \llm calls in a manner that yields complementary gains, i.e., the resulting performance exceeds that of any standalone model. In this sense, \methodname goes beyond routing, learning to orchestrate an \ahd process that leverages the collective strengths of multiple \llms.

\begin{figure}[t]
    \centering
    \vspace{-0.25in}
    \includegraphics[width=\linewidth]{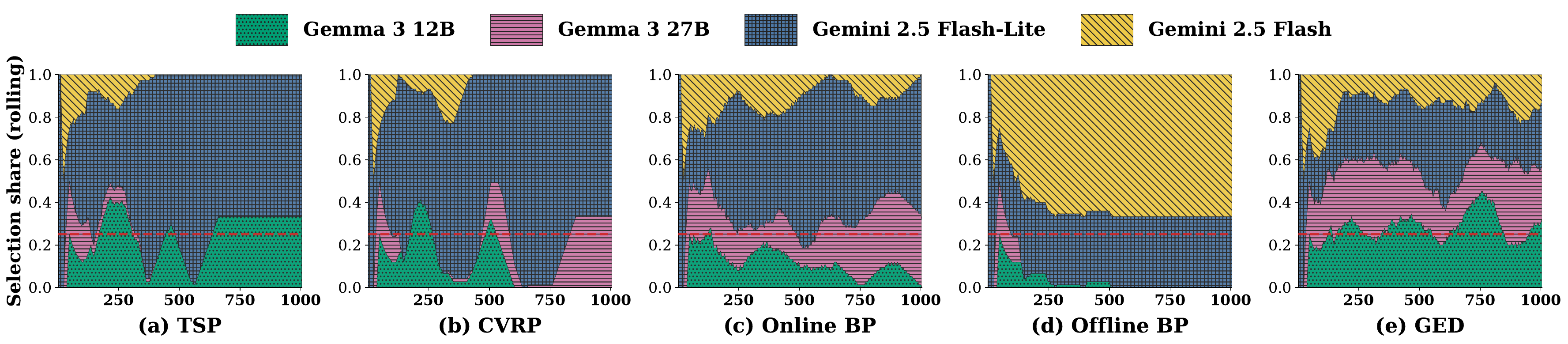}
    \caption{
        \textbf{LLM selection share across search steps (rolling window of 100 calls).} Each panel shows, for a combinatorial optimization problem, the fraction of LLM calls routed to each model by \methodname over the course of the evolutionary search. The dotted horizontal line marks the uniform-allocation baseline (1/K=25\% here), corresponding to a policy with no preference over arms. Deviations from this baseline reflect the learned routing policy by \methodname favors models with selection share above 1/K due to higher estimated utility. 
    }
    \label{fig:sequence}
    \vspace{-0.25in}
\end{figure}
To further understand how \methodname routes \llm calls, we analyze the sequence of selections over iterations $t \in [0,1000]$ in Fig.~\ref{fig:sequence}. Across most tasks, \methodname predominantly favors \flashlite, which is consistent with its strong average performance observed in Table~\ref{tab:test_comparison}. As iterations progress, the selection distribution becomes increasingly concentrated, with \methodname effectively discarding two models and exploiting the remaining ones. This behavior aligns with the expected exploration--exploitation dynamics of the bandit phase, where later rounds prioritize exploitation.

In the offline-BP problem, the final phase is dominated by \flash and \flashlite, which is consistent with their comparable performance in Table~\ref{tab:test_comparison}. An interesting deviation occurs for CVRP: although \gemmasmall achieves the best standalone performance, \methodname primarily selects \flashlite and \gemma. Despite not focusing on the individually best model, \methodname still produces heuristics that achieve the best objective and rank in CVRP (Table~\ref{tab:test_comparison}). This shows that high-quality solutions can emerge from the interaction and mutation of outputs generated by non-optimal models, highlighting the importance of credit distribution through the mutational lineage in the \ahd process.

In contrast, GED exhibits a different pattern, where all \llms remain competitive throughout the iterations. This indicates that no single model is consistently dominated, and the \ahd evolution maintains diversity along the performance frontier until the final stages.

We discuss the evolution of the best test objective over LLM calls, illustrating the progression and convergence of \methodname in App.~\ref{app:epoch_vs_perf}.

\begin{figure}[b!]
\vspace{-0.08in}
    \centering
    \includegraphics[width=\linewidth]{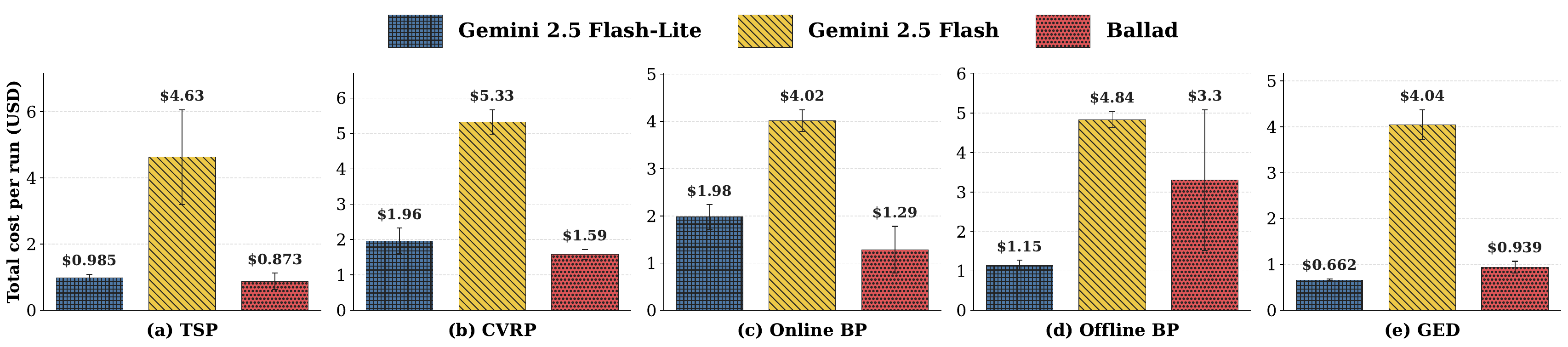}
    \vspace{-0.2in}
    \caption{
        \textbf{LLM API cost per run.}
        Total monetary cost (USD) of one evolutionary search run on each task, averaged over three independent runs (mean $\pm$ std as error bars).  Costs are
        computed from per-call input/output token counts using public Gemini~2.5 list prices: Flash-Lite at \$0.10/M input and \$0.40/M output tokens, Flash at \$0.30/M input and \$2.50/M output tokens. gemma-3 models incur no monetary cost, hence not shown. Average cost reduction relative to \flash using \methodname is 65.6\%.
    }
    \label{fig:cost_comparison}
    \vspace{-0.3in}
\end{figure}
\vspace{-0.1in}
\subsection{RQ3. Cost effectiveness.}
\label{sec: cost_effectiveness}
\vspace{-0.05in}
Fig.~\ref{fig:cost_comparison} reports the inference 
cost incurred by \methodname relative to \flash and 
\flashlite across all five problems. Against \flash, 
\methodname is up to $5.30\times$ cheaper and on 
average $2.86\times$ cheaper across all problems. 
Against \flashlite, the cheaper of the two paid 
models, \methodname is cheaper on three out of five 
problems, and only $1.19\times$ more expensive on 
average, despite exceeding \flashlite's 
performance across the board. The exceptions are 
Offline BP and GED; as Fig.~\ref{fig:sequence} shows, these are precisely the problems where \flash remains 
consistently competitive throughout the full horizon, 
leading the routing policy to sustain higher 
allocation to the strongest model.

These results highlight the core dilemma that 
\methodname resolves. Given a pool of \llms, there is 
no a priori way to determine which model is best 
suited for a given task. Discarding expensive models 
is risky, as stronger models may exhibit superior 
reasoning in some settings; however, as RQ1 shows, 
smaller and cheaper models frequently outperform 
them. The only reliable naive strategy is to evaluate 
all candidates exhaustively, which multiplies cost by 
the pool size. \methodname sidesteps this dilemma by 
automatically identifying an effective composition of 
\llms without exhaustive evaluation, not only 
matching but often surpassing the best individual 
model, while incurring only a fraction of the cost. 
Notably, this cost efficiency emerges implicitly: the 
index score (Eq.~\ref{eq:score}) is entirely agnostic 
to model cost, and the savings arise naturally from 
the empirical finding that the most expensive model 
is not consistently optimal. If desired, the 
framework can be readily extended to explicitly 
incorporate inference cost into the arm scoring 
function (Eq.~\ref{eq:score}).

\begin{table}[t]
\centering
\vspace{-0.35in}
\small
\caption{\textbf{Ablation Study: } The first four rows are baseline arm-selection strategies. $\lambda=0$ and $\lambda=1$ denote quantile- and maximum-based variants, respectively; $\alpha=0$ disables lineage-aware exploitation. $\tau$ controls the upper-tail level. \emph{Exp.\ bonus~$\propto\sqrt{\log t}$} uses standard \textsc{UCB} exploration. \emph{Cost-Penalized Exp.} accounts for \llm call costs and penalises exploration for expensive arms. \textbf{Bold} and \underline{underline} denote the best and second-best, respectively. Due to computational constraints, we evaluate the ablations on three problems.}
\label{tab:ablation}
\vspace{-0.005in}
\scalebox{0.92}{%
\begin{tabular}{llcccc}
\toprule
\textbf{Model} & \textbf{Metric} & \textbf{TSP} & \textbf{CVRP} & \textbf{Online BP} & \textbf{Overall} \\
\midrule
\textsc{Random} & Obj. & 6.243 $\pm$ 0.037  & 10.851 $\pm$ 0.087 & 2.071 $\pm$ 0.621 & 6.378 $\pm$ 0.363 \\
\textsc{UCB} & Obj. & \textbf{6.194 $\pm$ 0.051} & 10.896 $\pm$ 0.088 & 1.640 $\pm$ 0.616 & 6.243 $\pm$ 0.360 \\
\textsc{SW-UCB} & Obj. & 6.306 $\pm$ 0.072 & 10.885 $\pm$ 0.118 & 1.752 $\pm$ 0.087 & 6.314 $\pm$ 0.094 \\
\textsc{Thompson Sampling} & Obj. & 6.294 $\pm$ 0.004 &  10.861 $\pm$ 0.023 & 2.398 $\pm$ 0.112 & 6.517 $\pm$ 0.066 \\
\midrule
\methodname($\lambda=0$) & Obj. & 6.255 $\pm$ 0.043 & 10.869 $\pm$ 0.086 & 1.603 $\pm$ 0.162 & 6.242 $\pm$ 0.108 \\
\methodname($\lambda=1$) & Obj. & 6.250 $\pm$ 0.020 & 10.729 $\pm$ 0.017 & 1.988 $\pm$ 0.422 & 6.322 $\pm$ 0.244 \\
\midrule
\methodname{}($\alpha=0$) & Obj. & 6.281 $\pm$ 0.130 & 10.714 $\pm$ 0.334 & \textbf{1.483 $\pm$ 0.276} & \underline{6.159 $\pm$ 0.261} \\
\midrule
\methodname{}($\tau=0.7$) & Obj. & 6.444 $\pm$ 0.231  &  \textbf{10.625 $\pm$ 0.147} & 1.541 $\pm$ 0.447 & 6.203 $\pm$ 0.303 \\
\methodname{}($\tau=0.99$) & Obj. & 6.337 $\pm$ 0.013 &  10.695 $\pm$ 0.017 & 1.777 $\pm$ 0.733 & 6.270 $\pm$ 0.423 \\
\midrule
\methodname(Exp. Bonus $\propto \sqrt{\log t}$) & Obj. & 6.251 $\pm$ 0.024 & \underline{10.677 $\pm$ 0.069} & 2.124 $\pm$ 0.509 & 6.351 $\pm$ 0.297 \\
\methodname{}(Cost-Penalized Exp.) & Obj. & 6.266 $\pm$ 0.028 & 10.863 $\pm$ 0.135 & 2.079 $\pm$ 0.557 & 6.403 $\pm$ 0.331 \\
\midrule
\methodname & Obj. & \underline{6.233 $\pm$ 0.056} & 10.685 $\pm$ 0.075 & \underline{1.483 $\pm$ 0.276} & \textbf{6.134 $\pm$ 0.168} \\
\bottomrule
\end{tabular}}
\end{table}

\vspace{-0.1in}
\subsection{RQ4. Comparison with baselines and ablation of \methodname components}
\label{sec:ablations}
\vspace{-0.05in}
The ablation study (Table~\ref{tab:ablation}) isolates the contributions of \methodname’s core algorithmic components. By prioritizing upper-tail \textit{breakthroughs} over mean performance (P1), \methodname achieves a superior overall objective(6.134) to the standard \textsc{UCB}(6.243), \textsc{SW-UCB}(6.314) and Thompson Sampling(6.517) baselines (details in App. \ref{app: ucb}). The hybrid mixing of bootstrap quantiles and empirical maxima provides a balanced \textit{floor-and-ceiling} effect that outperforms both pure quantile-only and record-only configurations. Furthermore, disabling lineage-aware credit propagation ($\alpha=0$) degrades performance on TSP and CVRP, validating the causal coupling of models across the mutation genealogy (P2). The tail-level parameter $\tau$ balances emphasis on upper-tail performance against estimation reliability: lower values dilute breakthrough signals, shifting the estimate toward average-case performance, whereas higher values rely on fewer order statistics, yielding noisier estimates. Calibrating the exploration bonus to the fixed horizon $T$ rather than the current round $t$ concentrates search on fertile models and consistently outperforms growing-bonus schedules. Furthermore, cost-penalized exploration underperforms by favoring cheaper, weaker models over potentially breakthrough-generating arms.\\

We examine the impact of expanding the arm pool to nine models, including frontier models such as GPT-5 and Claude Sonnet 5, alongside Gemini 2.5 Flash, on Offline BP and GED in Table~\ref{tab:scaled_quality}. The expanded pool further improves objective values on both problems, while Table~\ref{tab:scaled_cost} reports cost savings averaging 46.35\%, with a maximum of 68.3\%, compared to always using GPT-5 under the assumption of comparable token usage. We provide a comprehensive analysis of the ablation results across all dimensions in App.~\ref{app:ablation_study}.\looseness=-1

%% file: conclusion.tex
\vspace{-0.10in}
\section{Conclusions, Limitations and Future Directions}
\label{sec:conclusion}
\vspace{-0.10in}
We presented \methodname, a bandit-based approach for adaptive \llm routing in automated heuristic discovery. We show that larger, more expensive models are not consistently superior, making exhaustive evaluation the only reliable but a prohibitively costly strategy. \methodname addresses this by learning to orchestrate a pool of \llms, leveraging extreme-value scoring, lineage-aware credit assignment, and non-stationary adaptation to identify effective model compositions. Across multiple combinatorial optimization problems, \methodname not only matches but often surpasses the best individual \llm, while significantly reducing inference cost. Notably, our results show that the optimal strategy is not to select a single model, but to dynamically compose multiple models based on their evolving utility.

\textbf{Limitations and Future Work}
Our current formulation treats each \llm as a black-box generator with a fixed prompting strategy. This overlooks that different \llms may benefit from customized prompts, mutation operators, incorporation of lineage information or context representations. 

\textbf{Impact Statement}
\label{sec:impact}
Potential risks are limited, as the method introduces no new models or datasets, but inherits general considerations of \llms, including environmental impact.
\vspace{-0.1in}
\section*{Acknowledgement}
\vspace{-0.05in}
We thank the Yardi School of Artificial Intelligence at the Indian Institute of Technology Delhi for sponsoring Samidha Verma’s expenses to attend NeurIPS 2026. We also thank NeurIPS for providing her with financial aid for accommodation and conference registration. We also acknowledge the Google PhD Fellowship awarded to Samidha Verma and the additional research funding provided by Incept Labs to support this work.

%% file: appendix.tex
\appendix
\input{method_app}
\section{Benchmark CO Problems and Datasets}
\label{app:problem_description}
In this section, we describe the NP-hard combinatorial optimization problems used to benchmark the efficacy of \methodname. Further, the dataset-specific splits and configurations are presented in Table~\ref{tab:dataset_comprehensive}.

\begin{table}[htbp]
\centering
\small
\setlength{\tabcolsep}{5pt}
\caption{%
\textbf{Benchmark dataset configurations.}
Problems are grouped by type. \textit{Problem Size} denotes the average instance size: number of cities/nodes for routing problems, number of items for packing problems, and $(\overline{|V|}, \overline{|E|})$ for graph problems. Problem-specific constraints, such as vehicle capacity $W$ (CVRP) and bin capacity $C$ (BP), are listed in the \textit{Constraints} column. Instance counts are given per split.
}
\label{tab:dataset_comprehensive}
\begin{tabular}{lccrrr}
\toprule
\textbf{Problem} & \textbf{Problem Size} & \textbf{Constraints}
& \textbf{Train} & \textbf{Val} & \textbf{Test} \\
\midrule
\multicolumn{6}{l}{\textit{Routing}} \\[1pt]
\quad TSP          & 50       & {-}                   & 64  & 64  & 1{,}000 \\
\quad CVRP         & 50       & $W = 50$              & 10  & 64  & 64 \\
\specialrule{0.08em}{0.3em}{0.3em}
\multicolumn{6}{l}{\textit{Bin Packing}} \\[1pt]
\quad Online BP   & 5{,}000  & $C \in \{100,\,500\}$ & 2   & 5   & 5 \\
\quad Offline BP  & 500      & $C = 150$             & 5   & 64  & 64 \\
\specialrule{0.08em}{0.3em}{0.3em}
\multicolumn{6}{l}{\textit{Graph Matching}} \\[1pt]
\quad GED (Linux)  & $(8, 7)$ & {-}                   & 1{,}000 & 1{,}000 & 1{,}000 \\
\bottomrule
\end{tabular}
\end{table}
\paragraph{Travelling Salesman Problem .} The Traveling Salesman Problem (TSP) is defined on a vertex set
$V = \{v_1, v_2, \ldots, v_N\}$ with distance matrix
$D = [d_{ij}] \in \mathbb{R}^{N \times N}$. 
A feasible solution is a Hamiltonian cycle (tour) $\mathcal{T} = (v_1, \ldots, v_N)$ visiting each vertex exactly once, with cost
\begin{equation}
    \minimise \; f_{\mathcal{P}}(\mathcal{T})
    = \sum_{t=1}^{N-1} d_{\mathcal{T}_t, \mathcal{T}_{t+1}} + d_{\mathcal{T}_N, \mathcal{T}_1}.
\end{equation}
In our setting, a heuristic $h$ maps an instance $I \in \mathcal{I}_{\mathcal{P}}$
to a tour $\mathcal{T} = h(I)$, which is evaluated by $f_{\mathcal{P}}(\mathcal{T})$.
Given a partial tour, the generated heuristic selects the next node to visit. We consider 64 problem instances with $N = 50$ nodes~\cite{zhengmonte}.

\paragraph{Capacitated Vehicle Routing Problem .} The Capacitated Vehicle Routing Problem (CVRP) is defined on a vertex set
$V = \{v_0, v_1, \ldots, v_N\}$, where $v_0$ denotes the depot and
$\{v_1,\ldots,v_N\}$ are customer nodes, together with a distance matrix
$D = [d_{ij}] \in \mathbb{R}^{(N+1)\times(N+1)}$, customer demands
$\{\delta_i\}_{i=1}^{N}$, and vehicle capacity $C$.
A feasible solution is a set of routes
$\mathcal{R} = \{\mathcal{T}^1, \ldots, \mathcal{T}^q\}$, where each route
$\mathcal{T}^j = (v_0, \mathcal{T}^j_1, \ldots, \mathcal{T}^j_{n_j}, v_0)$
starts and ends at the depot and visits a subset of customers.
The objective is to minimize the total travel cost,
\begin{equation}
    f_{\mathcal{P}}(\mathcal{R})
    =
    \sum_{j=1}^{q}
    \sum_{t=0}^{n_j}
    d_{\mathcal{T}^j_t,\; \mathcal{T}^j_{t+1}},
\end{equation}
subject to the capacity constraints
\begin{equation}
    \sum_{v_i \in \mathcal{T}^j} \delta_i \le C, 
    \quad \forall j \in \{1,\ldots,q\},
\end{equation}
and the requirement that each customer is visited exactly once.
In our setting, a heuristic $h$ parameterizes an Ant Colony Optimization (ACO) procedure, which constructs a solution $\mathcal{R} \sim \mathrm{ACO}(I; h)$ for an instance $I \in \mathcal{I}_{\mathcal{P}}$, and is evaluated via $f_{\mathcal{P}}(\mathcal{R})$. We utilize the dataset introduced in ~\cite{zhengmonte} where the  CVRP instances have the depot fixed at $(0.5, 0.5)$.

\paragraph{Bin Packing Problems .}The Bin Packing Problem (BPP) is a classical NP-hard combinatorial optimization problem defined over a set of items with sizes $\{s_i\}_{i=1}^{N}$ and bins of fixed capacity $W$. The objective is to assign each item to a bin such that the total size of items in any bin does not exceed $W$, while minimizing the total number of bins used. 

In the online setting, items arrive sequentially, and each assignment decision must be made irrevocably upon arrival, whereas in the offline setting, the full item set is available in advance. In our framework, a heuristic $h$ maps an instance $I \in \mathcal{I}_{\mathcal{P}}$ to a packing solution, which is evaluated via $f_{\mathcal{P}}(h(I))$, where $f_{\mathcal{P}}$ denotes the number of bins used. 

For the online setting, the heuristic is designed to extend a partial assignment step by step, selecting each next action based on the current state. In contrast, for offline bin packing, the heuristic is designed to construct a complete packing using an Ant Colony Optimization (ACO)--style procedure: items are processed sequentially, and each item is assigned to a bin by sampling from a probability distribution guided by pheromone values and feasibility information such as remaining bin capacity.

We utilize the dataset introduced in~\cite{zhengmonte} for our study. This setup is consistent with prior work in the literature, including~\cite{romera2024mathematical, liu2024evolution}.

\paragraph{Graph Edit Distance .}
Graph Edit Distance (GED) between two graphs $G_1=(V_1,E_1)$ and $G_2=(V_2,E_2)$ is defined as the minimum cost of transforming $G_1$ into $G_2$ through a sequence of edit operations, such as node and edge insertions, deletions, and substitutions. Formally, letting $\mathcal{M}$ denote the set of all possible node mappings between the two graphs, GED can be written as
\begin{equation}
\mathrm{GED}(G_1,G_2)=\min_{\pi \in \mathcal{M}} \mathrm{GED}_\pi(G_1,G_2),
\end{equation}
where $\mathrm{GED}_\pi(G_1,G_2)$ is the edit cost induced by a particular node mapping $\pi$. Since the number of possible mappings grows factorially with graph size, exact computation is NP-hard.

We follow \textsc{GRAIL}'s~\cite{verma2025grail} formulation for GED approximation, where the prompt includes the task specification and the top-$k$ programs discovered so far. Given a graph pair $(G_1,G_2)$, a heuristic $h$ generates a weight matrix $W$ defining a weighted bipartite graph; maximum-weight bipartite matching then produces a node mapping $\pi$, which is used to compute the resulting edit cost.

The cost induced by a mapping $\pi$ is given by
\begin{equation}
f_{\mathcal{P}}(h(G_1,G_2))
=
\sum_{v_1 \in V_1} \mathbb{I}\!\left(L_1(v_1) \neq L_2(\pi(v_1))\right)
+
\frac{1}{2}
\sum_{v_1 \in V_1}
\sum_{v_2 \in V_1}
\mathbb{I}\!\left(e_1(v_1,v_2) \neq e_2(\pi(v_1),\pi(v_2))\right),
\end{equation}
where $L(\cdot)$ denotes node labels and $e(\cdot,\cdot)$ indicates edge existence. The objective is to
\begin{equation}
\min_{h} \; f_{\mathcal{P}}(h(G_1,G_2)).
\end{equation}

While the original formulation targets a diverse set of complementary heuristics via a submodular objective, we instead use the same pipeline to learn a single heuristic that minimizes GED for each graph pair. 

We evaluate on the Linux benchmark from~\cite{verma2025grail}, which consists of 1,000 unlabeled program dependence graphs extracted from the Linux kernel, with nodes representing statements and edges capturing dependencies. The graphs are small (on average 8 nodes and 7 edges), making the dataset a standard testbed for GED heuristics. 

Problem-specific seed programs and the best programs on the validation set, discovered across all runs of \methodname have been provided in App.~\ref{sec:priority-v0} and App.~\ref{app:best_programs}, respectively.

\begin{figure}[htbp!]
\centering
\includegraphics[width=\textwidth]{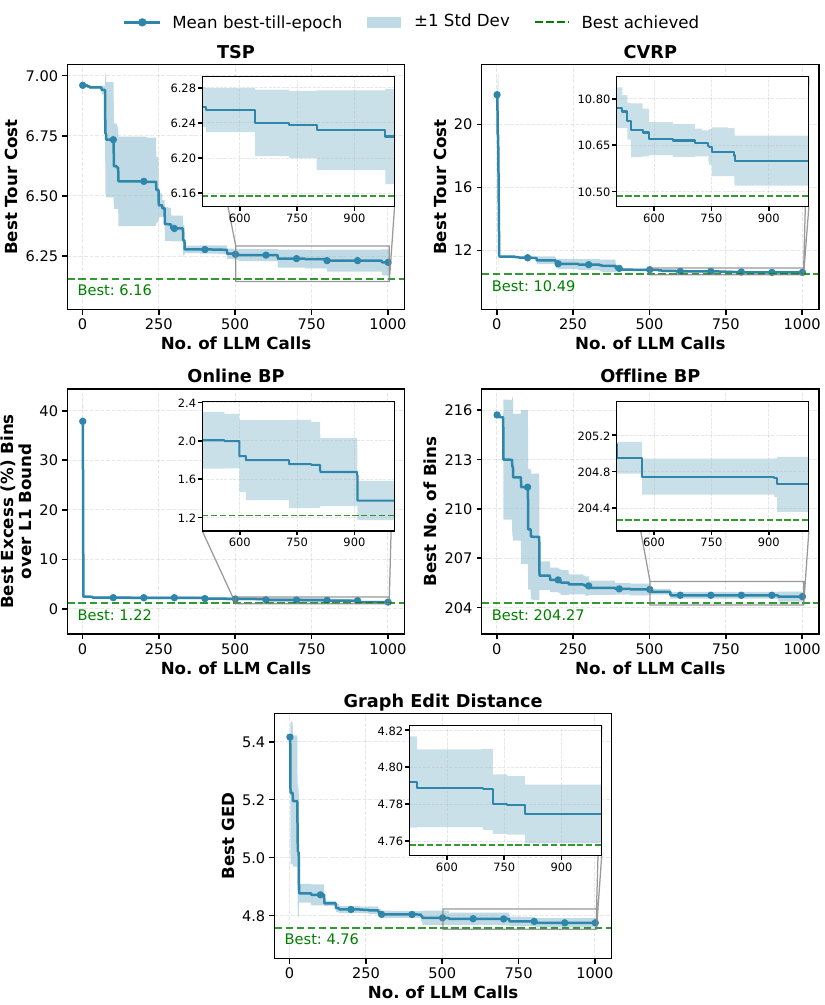}
\caption{\textbf{Best test objective vs.\ number of LLM calls for \methodname{} (lower is better).} After each call, all programs generated so far are evaluated on the test set, and the best-performing program is retained. The blue line shows the mean across three runs, with the shaded region indicating $\pm$1 standard deviation. The green dashed line marks the best objective achieved across all runs. Insets zoom in on the second half of the search to highlight smaller improvements. Since the best program is selected using test performance at each step, these curves provide an optimistic estimate of the results in Table~\ref{tab:test_comparison}, where selection is based on validation performance. The final mean is comparable to or better than the corresponding table value.}
\label{fig:epoch_vs_perf_all}
\end{figure}

\section{Experimental Setup: Details}
\label{app:reproducibility}
\subsection{Hyperparameters.}\label{app:params} In our experiments, we set $k=2$ (number of heuristics in the prompt) and follow the sampling strategy of~\cite{romera2024mathematical}, with a total budget of $T=1000$ evaluations. Hyperparameters are selected based on validation performance, with tuning restricted to the most sensitive ones ($\lambda$, $\alpha$, and $D_{\max}$), while others are fixed to reasonable defaults from preliminary experiments. The index score (Eq.~\ref{eq:score}) uses $\lambda = 0.3$, $\alpha = 1.0$, and $\beta = 1.0$. The lineage term is computed with a sliding window $W = 25$ rounds, discount factor $\gamma = 0.9$, and maximum depth $D_{\max} = 4$. The bootstrap estimator uses $B = 200$ resamples at quantile $\tau = 0.9$, and a warm-up threshold of $n_{\min} = 3$ observations per \llm. For getting diverse outputs, the temperature hyperparameter of \llms is set to $0.99$. A complete list of hyperparameters is provided in Table~\ref{tab:notations_v2}. Further, we ran the evolutionary search, i.e., each run for a maximum of 24 hours, during which a majority of the time was spent evaluating the generated heuristics. To handle throttling due to API rate limits, a 10-minute backoff with a maximum of 5 retries was also applied.

\subsection{Hardware Details.}\label{app:hardware}
Hardware details All experiments were performed on Intel(R) Xeon(R) Gold 6426Y: 64 cores, 126 GB RAM, 2 NVIDIA-L40S GPUs, 45 GiB each running on Ubuntu 20.04.6 LTS. The Gemini API was used to access \llms \flashlite and \flash under the paid tier, and \gemmasmall and \gemma under the free tier. The evaluations of TSP, Online BP, and GED were CPU-bound, whereas ACO-based CVRP and Offline BP were GPU-based. 


 
\section{LLM Calls vs. Test Performance.}
\label{app:epoch_vs_perf}
Figure~\ref{fig:epoch_vs_perf_all} shows how the best test objective of \methodname{} evolves with the number of LLM calls. Across all problems, most improvements occur early in the search, with the initial calls yielding substantial gains. As the best program becomes stronger, finding better candidates becomes increasingly difficult, and progress slows to occasional, smaller improvements (insets). Nevertheless, these late improvements remain meaningful, as they refine an already strong program. 



These curves provide an optimistic estimate because they report the best program on the test set, whereas Tables~\ref{tab:test_comparison} and ~\ref{tab:ablation} report the program selected using validation data. The final means of the curves are therefore at or below the table values (curve vs.\ table: TSP 6.225 vs.\ 6.233; CVRP 10.600 vs.\ 10.685; Online BP 1.375 vs.\ 1.483; Offline BP 204.661 vs.\ 204.693; GED 4.775 vs.\ 4.789). These differences reflect the generalization gap between selecting programs on validation data and evaluating them on the test set. Since all models in Tables~\ref{tab:test_comparison} and ~\ref{tab:ablation} follow the same validation-based selection procedure, the comparison between them remains consistent and fair.

\section{Discussion: Ablation Study}
\label{app:ablation_study}

We here expand upon the ablation study results from Table~\ref{tab:ablation}, offering a granular discussion of how individual design choices influence performance across various problem dimensions. This analysis clarifies the mechanistic contribution of each algorithmic component to the overall discovery process.

\paragraph{Impact of Extreme Value Objective.}
\label{sec: abl_evt}
Aligning with property [P1](Refer section~\ref{sec:methods}), \methodname optimizes for the upper-tail quality of the heuristic population rather than mean performance. This \textit{tail-hunting} strategy enables it to outperform the \textsc{UCB} baseline on the majority of tasks, achieving a statistically robust lead on CVRP($10.685 \pm 0.075$ vs. $10.896 \pm 0.088$)  where performance intervals show no overlap. While \textsc{UCB} exhibits a marginal edge on TSP($6.194 \pm 0.051$ vs. $6.233 \pm 0.056$) and \methodname shows a lower mean on Online BP($1.483 \pm 0.276$ vs. $1.640 \pm 0.616$), these specific differences are not statistically significant due to overlapping standard deviation intervals. Ultimately, \methodname delivers a superior aggregate objective value of 6.134 compared to \textsc{UCB}’s 6.243, establishing extreme-value orchestration as a more potent approach for \ahd.\\

The advantage also extends to more advanced mean-reward-based strategies. Both \textsc{SW-UCB} and \textsc{Thompson Sampling} yield worst objective values than \methodname across all three tasks, with aggregate values of 6.314 and 6.517, respectively, compared to 6.134. This indicates that adapting reward estimates to recent observations or incorporating posterior uncertainty alone does not capture the benefits of explicitly prioritizing the upper tail of heuristic quality. The detailed implementation of all baselines has been provided in App.~\ref{app: ucb}.

\paragraph{Impact of $\lambda$ (Max-Quantile Mixing).}
\label{sec: abl_lambda}
The parameter $\lambda$ balances the stability of the bootstrap quantile ($\lambda=0$) with the aggressive pursuit of empirical maximums ($\lambda=1$). Table~\ref{tab:ablation} confirms that neither extreme is optimal on its own. While the record-based approach ($\lambda=1$) outperforms the conservative quantile-only version ($\lambda=0$) on both TSP ($6.250 \pm 0.020$ vs. $6.255 \pm 0.043$) and CVRP ($10.729 \pm 0.017$ vs. $10.869 \pm 0.086$), it remains vulnerable to being "blinded" by non-replicable \llm outliers. The hybrid configuration used in \methodname yields the best overall objective ($6.134 \pm 0.168$) compared to the quantile-only baseline ($6.242 \pm 0.108$), demonstrating that the quantile provides a reliable quality "floor", while the empirical maximum provides the necessary "ceiling" to drive frontier discovery.  

\paragraph{Impact of Tail level $\tau$.}
\label{sec: abl_tau}
$\tau$ sets how far into the upper tail the base score reads in Eq.~\ref{eq:score}. \methodname{} uses $\tau=0.90$ by default. Both extremes degrade (Refer Table~\ref{tab:ablation}): at $\tau=0.70$ the estimator reverts toward average-case behaviour, at $\tau=0.99$ it rests on too few order statistics and becomes unstable (Online BP, ±0.733).

\paragraph{Impact of Lineage.}
\label{sec: abl_lineage}
Setting $\alpha = 0$ disables the propagation of rewards through the mutation genealogy, effectively treating each \llm as an isolated generator. This configuration results in a performance degradation across almost all metrics, including TSP ($6.281 \pm 0.130$ vs. $6.233 \pm 0.056$) and CVRP ($10.714 \pm 0.334$ vs. $10.685 \pm 0.075$). These results validate property [P2]: automated discovery is a combinatorial process where a "weaker" model may provide the foundational code necessary for a "stronger" model to achieve a breakthrough. While Online BP remains unchanged ($1.483$), the superior Overall objective of \methodname ($6.134$) compared to the lineage-blind version ($6.159$) confirms that rewarding the causal influence of ancestor models is essential for orchestrating an effective multi-model search.

To further examine the robustness of the impact of lineage, we evaluate Offline BP and GED using the nine-model pool variant, \methodname{} (Scaled, $\alpha=0$) in Table~\ref{tab:scaled_quality}. Lineage improves the objective on four of the five problems and ties on the fifth (Online BP, 1.483 in both settings). The comparison with the four-model configuration is more telling. With only four models, \methodname{} matches the nine-model pool without lineage on Offline BP (204.693 vs.\ 204.734) and surpasses it on GED (4.789 vs.\ 5.178, with non-overlapping intervals). Adding five frontier models therefore does not compensate for removing lineage credit.

\paragraph{Impact of Fixed Horizon Exploration Bonus.}
\label{sec: abl_exploration}
Standard \textsc{UCB} algorithms utilize an exploration bonus that grows with the current round $t$, which can lead to excessive exploration in the final stages of a fixed-budget search. By instead calibrating the bonus to the total horizon $T$, \methodname front-loads exploration when the budget is plentiful and shifts toward exploitation as the remaining budget depletes. As shown in Table~\ref{tab:ablation}, replacing this fixed-horizon schedule with the standard growing-bonus exploration ($\propto \sqrt{\log t}$) degrades overall performance, particularly on Online BP. This confirms that a budget-aware exploration schedule more effectively concentrates the final search effort on the most fertile models identified earlier in the run. 

\paragraph{Impact of Cost-Penalized Exploration.}
\label{sec: abl_cost_exploration}
Cost-penalised exploration probes the exploration term further, discounting each arm's bonus by its mean observed call cost via a softmax of the negative mean cost, i.e., $\rho(\ell)=e^{-\bar c_\ell}/\sum_{\ell'\in\text{paid}}e^{-\bar c_{\ell'}}$. Thus, the paid arms are discounted relative to one another in proportion to cost while free models (no cost observations) satisfy $\rho\equiv1$. 

Down-weighting exploration by price is worse on all three problems(Refer Table~\ref{tab:ablation}): it biases the search toward cheaper but weaker models, whereas in a discovery pipeline an arm's value lies in its capacity for rare breakthroughs. This substantiates section~\ref{sec: cost_effectiveness}; the 65.6\% saving is emergent, since an expensive model that underperforms simply accumulates few pulls.

\paragraph{Impact of Scaling to Frontier Models.}
\label{sec:abl_scaled}
We expand the arm pool from four to nine models: GPT-5, GPT-5 Mini, GPT-5 Nano, Claude Sonnet 5, Claude Haiku 4.5, Gemini 2.5 Flash, Gemini 2.5 Flash Lite, Gemma 3 27B, and Gemma 3 12B. We denote this variant \methodname{} (Scaled). Due to the prohibitive cost of running frontier models, we restrict this study to Offline BP and GED. As shown in Table~\ref{tab:scaled_quality}, the expanded pool improves quality on both problems. On GED, the standard-deviation intervals do not overlap, indicating a stronger improvement.

\begin{table}[t]
\centering
\vspace{-0.25in}
\small
\caption{\textbf{Scaling to Frontier Models:} Objective values on Offline BP and GED for four- and nine-model pools, with and without lineage-aware exploitation. \emph{Scaled} denotes the nine-model pool. \textbf{Bold} and \underline{underline} denote the best and runner-up values, respectively.}
\label{tab:scaled_quality}
\scalebox{0.85}{%
\begin{tabular}{lcc}
\toprule
\textbf{Model} & \textbf{Offline BP} & \textbf{GED} \\
\midrule
\methodname{} (Scaled, $\alpha=0$) & 204.734 $\pm$ 0.109 & 5.178 $\pm$ 0.238 \\
\methodname{} (Scaled) & \textbf{204.539 $\pm$ 0.242} & \textbf{4.706 $\pm$ 0.026} \\
\midrule
\methodname{} (Table~\ref{tab:test_comparison}, 4 models) & \underline{204.693 $\pm$ 0.323} & \underline{4.789 $\pm$ 0.022} \\
\bottomrule
\end{tabular}}
\end{table}

We analyze the cost of \methodname{} (Scaled) by computing the total input and output tokens over the runs of \methodname{}, pricing every call at the rate of the model that served it and averaging across runs. We compare this against pricing the identical token volume entirely at GPT-5's rate. This comparison assumes that GPT-5 alone would consume a comparable number of tokens, an assumption we do not verify empirically. Table~\ref{tab:scaled_cost} reports the cost per run. The saving is larger on GED because the router concentrates on cheaper arms there, whereas Offline BP sustains a higher allocation to strong models, consistent with the pattern reported in Section~\ref{sec: cost_effectiveness}.

\begin{table}[t]
\centering
\small
\caption{\textbf{Cost Comparison:} Average cost per run (USD) of \methodname{} (Scaled) versus pricing the same token volume entirely at GPT-5's rate. Relative savings are shown in parentheses.}
\label{tab:scaled_cost}
\scalebox{0.85}{%
\begin{tabular}{lccc}
\toprule
\textbf{Dataset} & \textbf{\methodname{} (Scaled)} & \textbf{GPT-5 only} & \textbf{Savings (USD)} \\
\midrule
Offline BP & 22.629 & 29.948 & 7.319 (24.4\%) \\
GED        & 9.745  & 30.739 & 20.994 (68.3\%) \\
\bottomrule
\end{tabular}}
\end{table}

\begin{algorithm}[htbp]
\caption{UCB-based \llm Router}
\label{alg:llm_router_ucb1}
\begin{algorithmic}[1]
\setlength{\itemsep}{1.5pt}
\Require 
$\begin{aligned}
&\text{\llms } \mathcal{L} = \{l_1, \ldots, l_n\},\quad
\text{instances } \mathcal{I}_{\mathcal{P}}, \\
&\text{horizon } T,\quad
\text{batch size } m,\quad
\text{exploration } \beta, \\
&\text{elite pool size } S
\end{aligned}$
\State Initialize $n_l \gets 0$,\; $\bar{U}_l \gets 0$ for all $l \in \mathcal{L}$
\State Initialize elite pool $\mathcal{E} \gets \emptyset$
\State $B \gets -\infty$

\For{$t = 1, 2, \dots, T$}
  \State \textbf{Router selection:}
  \State $l_t \gets \arg\max_{l \in \mathcal{L}}
      \bigl(\bar{U}_l + \beta \sqrt{\tfrac{\log t}{n_l + 1}}\bigr)$

  \State $\mathcal{U} \gets \emptyset$

  \For{$j = 1$ to $m$}
      \State Sample $h_t^{(j)} \sim \mu(l_t, H_t)$
      \State $S_t^{(j)} \gets \frac{1}{|\mathcal{I}_{\mathcal{P}}|}
      \sum_{I \in \mathcal{I}_{\mathcal{P}}}
      f_{\mathcal{P}}(h_t^{(j)}, I)$

      \State Update $\mathcal{E}$ with $(h_t^{(j)}, S_t^{(j)})$ (keep top-$S$)

      \If{$|\mathcal{E}| < S$}
          \State $B \gets \max(B, S_t^{(j)})$
      \Else
          \State $B \gets \mathrm{median}\bigl(\{s : (h,s) \in \mathcal{E}\}\bigr)$
      \EndIf

      \State $R_t^{(j)} \gets \max(0,\, S_t^{(j)} - B)$
      \State Append $R_t^{(j)}$ to $\mathcal{U}$
  \EndFor

  \State Aggregate round utility:
  \[
      U_t \;\gets\; \frac{1}{m} \sum_{u \in \mathcal{U}} u
  \]

  \State Update arm statistics:
  \[
      n_{l_t} \;\gets\; n_{l_t} + 1, \qquad
      \bar{U}_{l_t} \;\gets\;
      \bar{U}_{l_t} + \frac{U_t - \bar{U}_{l_t}}{n_{l_t}}
  \]

\EndFor
\end{algorithmic}
\end{algorithm}

\section{\llm Routing Baselines}
\label{app: ucb}
\subsection{Random}
The random router selects $l_t$ uniformly at random from $\mathcal{L}$ at every round, independent of past outcomes. 

\begin{algorithm}[htbp]
\caption{Thompson Sampling \llm Router}
\label{alg:llm_router_ts}
\begin{algorithmic}[1]
\setlength{\itemsep}{1.5pt}
\Require
$\begin{aligned}
&\text{\llms\ } \mathcal{L},\quad \text{instances } \mathcal{I}_{\mathcal{P}},\quad
\text{horizon } T,\quad \text{batch size } m, \\
&\text{elite pool size } S,\quad \text{prior } (\mu_0, \kappa_0, a_0, b_0),\quad \text{warm-up pulls } n_{\min}
\end{aligned}$
\State $(\mu_l, \kappa_l, a_l, b_l) \gets (\mu_0, \kappa_0, a_0, b_0)$,\; $n_l \gets 0$ for all $l \in \mathcal{L}$
\State Initialize elite pool $\mathcal{E} \gets \emptyset$,\; $B \gets -\infty$
\For{$t = 1, \dots, T$}
  \If{$t \le n_{\min} |\mathcal{L}|$}
    \State $l_t \gets l_{\lceil t / n_{\min} \rceil}$ \Comment{warm-up}
  \Else
    \For{each $l \in \mathcal{L}$}
      \State $\tilde{\sigma}_l^2 \sim \mathrm{InvGamma}(a_l, b_l)$,\quad
             $\tilde{\theta}_l \sim \mathcal{N}(\mu_l,\, \tilde{\sigma}_l^2 / \kappa_l)$
    \EndFor
    \State $l_t \gets \arg\max_{l} \tilde{\theta}_l$
  \EndIf
  \State Sample $m$ heuristics from $l_t$, evaluate on $\mathcal{I}_{\mathcal{P}}$, and compute
         $R_t^{(j)} = \max(0, S_t^{(j)} - B)$ as in Alg.~\ref{alg:llm_router_ucb1}
  \State Update $\mathcal{E}$ and $B$
  \State $\bar{R}_t \gets \frac{1}{m}\sum_{j=1}^{m} R_t^{(j)}, \quad
        Q_t \gets \sum_{j=1}^{m} \bigl(R_t^{(j)} - \bar{R}_t\bigr)^2$
\State $b_{l_t} \gets b_{l_t} + \frac{1}{2} Q_t
        + \frac{\kappa_{l_t}\, m\, (\bar{R}_t - \mu_{l_t})^2}{2(\kappa_{l_t} + m)}$
\State $\mu_{l_t} \gets \frac{\kappa_{l_t} \mu_{l_t} + m \bar{R}_t}{\kappa_{l_t} + m}$
\State $\kappa_{l_t} \gets \kappa_{l_t} + m, \quad a_{l_t} \gets a_{l_t} + \frac{m}{2}$
  \State $n_{l_t} \gets n_{l_t} + 1$
\EndFor
\end{algorithmic}
\end{algorithm}

\subsection{UCB}
We describe the UCB-based \llm routing baseline. At each round $t$, an arm $l_t \in \mathcal{L}$ is selected by maximizing the UCB index, and $m$ heuristics are sampled and evaluated on $\mathcal{I}_{\mathcal{P}}$. To define the reward baseline, we maintain an elite pool of the top-$S$ scoring programs observed so far. During the warm-up phase, when the pool contains fewer than $S$ elements, the baseline $B$ is set to the running maximum score. Once the pool is full, $B$ is defined as the median score of the elite set. The utility signal for arm $l_t$ is the mean relative improvement over this baseline,
\[
    R_t^{(j)} \;=\; \max\!\left(0,\; S_t^{(j)} - B\right),
\]
which credits an arm only when it advances the current elite frontier rather than rewarding absolute score levels. Using raw scores $S_t^{(j)}$ directly would penalize high-variance arms whose outputs are inconsistent but capable of rare, large improvements, precisely the arms most valuable in \ahd.
The baseline-relative reward may become sparse in practice, because once the search plateaus, most samples yield $R_t^{(j)}=0$. This makes it difficult for any mean-based estimator to discriminate between arms in later rounds; however, exploration helps escape such plateaus.

For exploration, UCB uses a bonus $\beta\sqrt{\log t\,/\,(n_l+1)}$ that grows with the elapsed round $t$, reflecting cumulative-regret guarantees in the classical bandit setting. In \ahd, however, the
horizon $T$ is fixed, and the objective is a single best outcome rather than a cumulative sum, so a growing bonus over-explores in later rounds when the remaining budget is scarce. The algorithm used has been provided in Alg.~\ref{alg:llm_router_ucb1}.

\subsection{SW-UCB}
Sliding-Window UCB~\citep{garivier2011upper} targets non-stationarity. The usefulness of an \llm may drift as the population evolves; for example, a model that contributes early can stop producing improvements once the search moves past the regime it is good at. Plain UCB responds to such drift slowly, because it averages over all time. SW-UCB is identical to UCB except that arm statistics are computed only over the most recent $W$ rounds.

Let $n_l^W(t)$ be the number of rounds among the last $W$ in which $l$ was selected, and let $\bar{U}_l^W(t)$ be the mean round utility $U_\tau$ over those rounds. The router then selects
\[
    l_t \;=\; \arg\max_{l \in \mathcal{L}}
    \left( \bar{U}_l^W(t) + \beta \sqrt{\frac{\log \min(t, W)}{n_l^W(t)}} \right).
\]
Any arm with $n_l^W(t) = 0$ receives an infinite index, so an arm that has not been pulled within the window is forced to be re-tried. Rewards, the elite-pool baseline, and the round utility $U_t$ are computed exactly as in Alg.~\ref{alg:llm_router_ucb1}. Capping the logarithm at $\log W$ also keeps the exploration bonus bounded. This partially mitigates the over-exploration of a growing bonus in late rounds noted above.

\subsection{Thompson Sampling}
Thompson Sampling~\citep{thompson1933likelihood} replaces the explicit exploration bonus with posterior sampling. Each individual reward from arm $l$ is modelled as $R \sim \mathcal{N}(\theta_l, \sigma_l^2)$, where both the mean and the variance are unknown. Each arm carries a conjugate Normal-Inverse-Gamma posterior $\mathrm{NIG}(\mu_l, \kappa_l, a_l, b_l)$, initialised at a shared prior $(\mu_0, \kappa_0, a_0, b_0)$. Since rewards are clipped at zero, the Gaussian likelihood is a working approximation. Its purpose is to provide tractable conjugate updates, not to be an exact model of the reward distribution.

At each round, one draw of the mean is taken from every arm's posterior, and the arm with the largest draw is pulled. Arms with wide posteriors are thus sampled optimistically (exploration), while arms with tight, high-mean posteriors are pulled more often (exploitation). After the round, the posterior of $l_t$ is updated in closed form with its $m$ rewards. Unlike UCB, which aggregates each round into a single utility $U_t$, the posterior is updated with all $m$ individual rewards. Confidence in an arm therefore grows with the number of evaluated heuristics rather than with the number of rounds.

Before sampling drives selection, a warm-up phase pulls each arm $n_{\min}$ times in sequence, so that every posterior has been updated at least once. We use $(\mu_0, \kappa_0, a_0, b_0) = (0, 1, 2, 1)$ and $n_{\min} = 1$. The full procedure, including the posterior update, is given in Alg.~\ref{alg:llm_router_ts}.

\section{Seed Heuristics:}
\label{sec:priority-v0}

For each combinatorial optimization problem, FunSearch~\citep{romera2024mathematical} begins with a seed program annotated by \texttt{@funsearch.evolve}. The function
below, \texttt{priority\_v0}, is the initial, deliberately simple,  heuristic that the \llm is asked to evolve. Its docstring defines the \emph{contract} (inputs, outputs, problem semantics) that every evolved variant must respect.

\subsection{Travelling Salesman Problem (TSP, constructive)}
\textbf{Idea.} Greedy nearest-neighbour: among unvisited nodes,
pick the one closest to the current node.

\begin{lstlisting}[style=pyseed,caption={TSP --- \texttt{priority\_v0}}]
@funsearch.evolve
def priority(current_node: int, destination_node: int,
             unvisited_nodes: set, distance_matrix: np.ndarray) -> int:
  """
    Selects the next node to visit in the Travelling Salesman Problem 
    (TSP) tour.

    The TSP aims to find a minimum-cost tour that visits each node exactly 
    once and returns to the starting node. This `priority' function 
    defines a heuristic for choosing the next node during the construction 
    of the tour.

    Args:
        current_node : int
            The index of the node currently being visited.
        destination_node : int
            The index of the tour's starting node (used for returning at the end of the tour).
        unvisited_nodes : set[int]
            The set of node indices that have not yet been visited.
        distance_matrix : np.ndarray
            A symmetric 2D array of shape (n, n) where entry [i, j] represents
            the Euclidean distance between node `i` and node `j`.

    Returns:
        int
            The index of the next node to visit, selected from `unvisited_nodes`
            according to the heuristic used in this function.
  """
  scores = {}
  for node in unvisited_nodes:
      scores[node] = distance_matrix[current_node, node]
  next_node = min(scores, key=scores.get)
  return next_node
\end{lstlisting}

\subsection{Capacitated Vehicle Routing Problem (CVRP (ACO))}
\textbf{Idea.} Inverse-distance heuristic: nearby edges are more
attractive. Demand and capacity are ignored at v0.

\begin{lstlisting}[style=pyseed,caption={CVRP --- \texttt{priority\_v0}}]
@funsearch.evolve
def priority(distance_matrix: np.ndarray, coordinates: np.ndarray,
             demands: np.ndarray, capacity: int) -> np.ndarray:
  """
  Computes prior indicators of how promising it is to include each edge 
  in a solution for the Capacitated Vehicle Routing Problem (CVRP).

  The CVRP extends TSP by using multiple vehicles with limited capacity 
  to serve customers with known demands, minimizing total travel distance 
  while respecting capacity constraints. The depot node is indexed by 0 
  and serves as the starting and ending point for all vehicle routes.

  Args:
      distance_matrix : np.ndarray
          2D array of distances between all node pairs (shape: n by n).
      coordinates : np.ndarray
          2D array of Euclidean coordinates of nodes (shape: n by 2).
      demands : list or np.ndarray
          Vector of customer demand requirements (shape: n). The depot node
          (index 0) has demand = 0.
      capacity : int or float
          Integer capacity constraint representing the maximum capacity
          of each vehicle.

  Returns:
      np.ndarray
          Heuristic matrix of the same shape as the distance_matrix, where
          entry [i, j] represents prior indicators of how promising it is
          to move from node i to node j, considering distance, demand,
          and capacity constraints.
  """
  return 1 / distance_matrix
\end{lstlisting}

\subsection{Online Bin Packing}
\textbf{Idea.} A handcrafted polynomial of remaining capacity and
item size, with a sign flip for bins that can fit the item and a
neighbour-difference smoothing pass.

\begin{lstlisting}[style=pyseed,caption={Online BPP --- \texttt{priority\_v0}}]
@funsearch.evolve
def priority(item: float, bins_remain_cap: np.ndarray) -> np.ndarray:
  """
    Computes a priority score for each bin in an Online Bin Packing problem.

    In the Online Bin Packing problem, items arrive sequentially and 
    must be placed into bins with limited capacity. This `priority` 
    function assigns a heuristic score to each bin, indicating how 
    suitable it is for accommodating the current item. The bin with the 
    highest priority score is selected for placement.

    Args:
        item : float
            The size of the current item to be packed.
        bins_remain_cap : np.ndarray
            A 1D array where each element represents the remaining capacity
            of a corresponding bin.

    Returns:
        np.ndarray
            A 1D array of priority scores for all bins. Each score reflects the
            relative suitability of placing the current item in that bin.
            The bin with the highest score is considered the best choice.
  """
  max_bin_cap = max(bins_remain_cap)
  score = (bins_remain_cap - max_bin_cap)**2 / item + bins_remain_cap**2 / (item**2)
  score += bins_remain_cap**2 / item**3
  score[bins_remain_cap > item] = -score[bins_remain_cap > item]
  score[1:] -= score[:-1]
  return score
\end{lstlisting}

\subsection{Offline Bin Packing (ACO)}
\textbf{Idea.} Pairwise heuristic matrix: each row is the demand
vector normalised by its maximum; capacity feasibility is left for
evolution to discover.

\begin{lstlisting}[style=pyseed,caption={Offline BPP --- \texttt{priority\_v0}}]
@funsearch.evolve
def priority(demand: np.ndarray, capacity: int) -> np.ndarray:
  """
  Computes a heuristic matrix that represents the relative suitability of
  grouping items together in the same bin for a bin packing problem.

  The output is a square matrix where each entry [i, j] indicates how
  promising it is to place item i and item j in the same bin, considering
  their demand values. Pairs (i, j) whose combined demand exceeds the bin 
  capacity should be assigned a lower heuristic score, as they cannot 
  feasibly be placed in the same bin.

  Args:
      demand : np.ndarray
          1D array of demand values for n items (shape: (n,)).
      capacity : int
          Integer specifying the maximum capacity of each bin. This provides
          an upper limit that should guide which item pairs are practically
          feasible to be grouped together.

  Returns:
      np.ndarray
          2D heuristic matrix of shape (n, n).
          Entry [i, j] signifies the prior indicator of how suitable it is
          to assign item i and item j to the same bin, taking demand
          compatibility and capacity feasibility into account.
  """
  return np.tile(demand / demand.max(), (demand.shape[0], 1))
\end{lstlisting}


\subsection{Graph Edit Distance (GED)}
\textbf{Idea.} A trivial seed: discard the input weights and return
a zero matrix. Evolution is expected to discover how to refine the
node-mapping probabilities using both adjacency matrices.

\begin{lstlisting}[style=pyseed,caption={GED --- \texttt{priority\_v0}}]
@funsearch.evolve
def priority(graph1: list[list[int]], graph2: list[list[int]],
             weights: list[list[int]]) -> list[list[float]]:
  """
    Computes the Graph Edit Distance (GED), a measure of the dissimilarity 
    between two graphs.
    
    GED is defined as the minimum number of operations required to 
    transform one graph into another. The primary operations considered 
    in GED calculations include:

    - **Node Insertion/Deletion:** Adding or removing a node incurs a 
      cost of +1.
    - **Edge Insertion/Deletion:** Adding or removing an edge between two nodes incurs a cost of +1.
    - **Node Relabeling:** Modifying the label of a node (if labels are present) adds a cost of +1 for each mismatch.

    Args:
        graph1: The adjacency matrix of the first graph.
        graph2: The adjacency matrix of the second graph.
        weights: A weight matrix representing the initial probabilities of mapping nodes           between `graph1` and `graph2`.
                 Each entry is a probability value, where a higher value indicates a higher likelihood and similarity of mapping nodes. The size of the weight matrix is determined by the maximum number of nodes in both graphs squared.

    Returns:
        A refined weight matrix (float) using the initial input matrix and the adjacency matrices of graphs where each entry represents the probability of a node in `graph1`
        being mapped to a node in `graph2` in a way that minimizes the overall graph edit distance.
  """
  max_node = len(graph1)
  weights = [[0.0] * max_node for _ in range(max_node)]
  return weights
\end{lstlisting}

\section{Best Discovered Heuristics}
\label{app:best_programs}

\subsection{Travelling Salesman Problem (TSP)}
\textbf{Idea.} Next-node selection via a weighted sum of immediate edge cost
and an estimated future path cost computed by a greedy nearest-neighbour
lookahead; the weight on immediate cost increases and the weight on future
cost decreases as the tour progresses, transitioning from a lookahead regime
at the start to a myopic nearest-neighbour regime near completion.

\begin{lstlisting}[style=pyseed,caption={TSP (Constructive) --- \texttt{priority\_v2}}]
@funsearch.evolve
def priority_v2(current_node: int, destination_node: int, unvisited_nodes: set, distance_matrix: np.ndarray) -> int:
  """
  Selects the next node to visit in the Travelling Salesman Problem (TSP) tour,
  prioritizing nodes that lead to a shorter estimated remaining tour, with a
  bias towards the immediate nearest neighbor when few nodes remain.

  This version is an improvement on `priority_v1` by more explicitly
  incorporating the total tour cost estimation into the priority calculation.
  It also uses a simpler, more robust dynamic weighting scheme.

  The logic:
  1. If only one node is unvisited, return it.
  2. For each `potential_next_node` in `unvisited_nodes`:
     a. Calculate the direct cost from `current_node` to `potential_next_node` (immediate cost).
     b. Estimate the "future path cost" from `potential_next_node` to the `destination_node`.
        This estimation is done by constructing a *temporary* greedy tour from
        `potential_next_node` through all other `unvisited_nodes - {potential_next_node}`,
        and finally back to `destination_node`.
     c. Dynamically determine weights for immediate and future costs:
        - As the number of unvisited nodes decreases (i.e., the tour progresses),
          the `w_immediate_cost` increases, and `w_future_path_cost` decreases.
        - This aims to make the heuristic behave more like a simple Nearest Neighbor
          at the end of the tour, and more like a lookahead-based heuristic at the beginning.
     d. Calculate the priority score as a weighted sum of immediate and future costs.
     e. Select the `potential_next_node` with the minimum priority score.

  Args:
      current_node : int
          The index of the node currently being visited.
      destination_node : int
          The index of the tour's starting node (used for returning at the end of the tour).
      unvisited_nodes : set[int]
          The set of node indices that have not yet been visited.
      distance_matrix : np.ndarray
          A symmetric 2D array of shape (n, n) where entry [i, j] represents
          the Euclidean distance between node `i` and node `j`.

  Returns:
      int
          The index of the next node to visit, selected from `unvisited_nodes`
          according to the heuristic used in this function.
  """
  num_unvisited = len(unvisited_nodes)
  total_nodes = distance_matrix.shape[0]

  if num_unvisited == 1:
    return list(unvisited_nodes)[0]

  best_next_node = -1
  min_priority_score = float('inf')

  # --- Dynamic Weighting Strategy ---
  # The goal is to shift from a broader lookahead to a more myopic (nearest neighbor)
  # focus as the tour progresses.
  #
  # Let's define a "progress" metric. A simple way is to consider the fraction of
  # nodes remaining to be visited *after* the next step.
  # If num_unvisited = N-1 (start), progress is near 0.
  # If num_unvisited = 1 (end), progress is near 1.
  #
  # We can use this progress to adjust weights.
  # `progress_factor` will range from ~0 (at start) to ~1 (at end).
  # Let's use `(total_nodes - 1 - num_unvisited) / (total_nodes - 1)` as the progress factor.
  # However, if `total_nodes` is 1 or 2, the denominator can be zero.
  # A simpler and more robust way: use `num_unvisited` directly.
  #
  # When `num_unvisited` is large, `w_future` should be higher and `w_immediate` lower.
  # When `num_unvisited` is small, `w_immediate` should be higher and `w_future` lower.

  # Define base weights
  base_immediate_weight = 1.0
  base_future_weight = 0.8

  # Calculate an adjustment factor based on the number of unvisited nodes.
  # This factor will be higher when `num_unvisited` is small.
  # Let's scale `num_unvisited` relative to `total_nodes`.
  # A factor from 0 to 1, where 1 means `num_unvisited` is 1, and 0 means `num_unvisited` is `total_nodes - 1`.
  # This makes `adjustment` closer to 1 at the end of the tour, and closer to 0 at the start.

  if total_nodes <= 2: # No dynamic adjustment needed for trivial cases
      adjustment = 0.0
  else:
      # `num_unvisited` goes from `total_nodes - 1` down to `1`.
      # We want `adjustment` to be close to 1 when `num_unvisited` is 1.
      # We want `adjustment` to be close to 0 when `num_unvisited` is `total_nodes - 1`.
      # Linear mapping: `adjustment = (num_unvisited - 1) / (total_nodes - 2)`
      adjustment = (num_unvisited - 1) / (total_nodes - 2)

  # Adjust weights:
  # As `adjustment` increases (tour progresses), increase `w_immediate` and decrease `w_future`.
  w_immediate_cost = base_immediate_weight + adjustment * 0.8  # e.g., increases from 1.0 to 1.8
  w_future_path_cost = base_future_weight * (1 - adjustment * 0.6) # e.g., decreases from 0.8 to 0.8 * (1-0.6) = 0.32

  for potential_next_node in unvisited_nodes:
    # a. Calculate the direct cost from `current_node` to `potential_next_node`.
    immediate_cost = distance_matrix[current_node, potential_next_node]

    # b. Estimate the "future path cost" from `potential_next_node` to the `destination_node`.
    # This is a full greedy Nearest Neighbor tour starting from `potential_next_node`.
    future_path_cost = 0

    remaining_unvisited_for_future = unvisited_nodes - {potential_next_node}

    if not remaining_unvisited_for_future:
        # If this is the last unvisited node, the future path is just to the destination.
        future_path_cost = distance_matrix[potential_next_node, destination_node]
    else:
        # Perform a greedy Nearest Neighbor tour from `potential_next_node` through all remaining nodes.
        temp_current_for_future = potential_next_node
        temp_unvisited_for_future = copy(remaining_unvisited_for_future)

        while temp_unvisited_for_future:
            closest_remaining_node_in_future = -1
            min_dist_to_remaining_in_future = float('inf')

            for node in temp_unvisited_for_future:
                dist = distance_matrix[temp_current_for_future, node]
                if dist < min_dist_to_remaining_in_future:
                    min_dist_to_remaining_in_future = dist
                    closest_remaining_node_in_future = node

            future_path_cost += min_dist_to_remaining_in_future
            temp_current_for_future = closest_remaining_node_in_future
            temp_unvisited_for_future.remove(closest_remaining_node_in_future)

        # Add the distance from the last visited node in the greedy path to the destination.
        future_path_cost += distance_matrix[temp_current_for_future, destination_node]

    # c. Calculate the priority score using dynamic weights.
    priority_score = (w_immediate_cost * immediate_cost) + \
                     (w_future_path_cost * future_path_cost)

    # d. Select the `potential_next_node` with the minimum priority score.
    if priority_score < min_priority_score:
      min_priority_score = priority_score
      best_next_node = potential_next_node

  return best_next_node
\end{lstlisting}


\subsection{Capacitated Vehicle Routing Problem (ACO)}
\textbf{Idea.} Edge heuristic matrix formed as a product of powered factors:
inverse distance, destination-demand penalty, source-demand preference,
destination peripheralness, source centrality, depot-return incentive, and
a customer reachability score; edges involving customers whose demand exceeds
vehicle capacity are zeroed and a small floor prevents stagnation.

\begin{lstlisting}[style=pyseed,caption={CVRP --- \texttt{priority\_v1}}]
@funsearch.evolve
def priority_v1(distance_matrix: np.ndarray, coordinates: np.ndarray, demands: np.ndarray, capacity: int) -> np.ndarray:
  """
  An improved heuristic for CVRP edge selection, building upon `priority_v0`.

  This version aims to refine the heuristic by:
  1.  **More Sophisticated Demand Handling:** Consider not just individual demand,
      but the *cumulative* demand on a potential route segment, relative to capacity.
      This is challenging to do purely statically. We'll approximate by considering
      the demand of the *source* node more strongly and how it contributes to a partial route load.
  2.  **Dynamic Cost Adjustment (Static Approximation):** Make edge costs
      more sensitive to how they fit into a potential route. For example,
      edges that lead to a high concentration of remaining demand might be
      less preferred unless they are close to the depot.
  3.  **Customer "Reachability" Score:** Nodes that are "harder" to reach (further
      from clusters, higher demand) might be prioritized earlier in a route
      to ensure they are not missed.
  4.  **Balancing Inter-vehicle Load:** While a static heuristic can't directly
      balance loads between vehicles, it can encourage choices that *tend* to
      lead to more balanced routes (e.g., by favoring shorter routes or
      routes that visit nodes with intermediate demands).
  5.  **Refined Peripheralness/Centrality:** Differentiate between peripheralness
      for the source and destination nodes.

  Args:
      distance_matrix : np.ndarray
          2D array of distances between all node pairs (shape: n by n).
      coordinates : np.ndarray
          2D array of Euclidean coordinates of nodes (shape: n by 2).
      demands : list or np.ndarray
          Vector of customer demand requirements (shape: n). The depot node
          (index 0) has demand = 0.
      capacity : int or float
          Integer capacity constraint representing the maximum capacity
          of each vehicle.

  Returns:
      np.ndarray
          Heuristic matrix of the same shape as the distance_matrix, where
          entry [i, j] represents prior indicators of how promising it is
          to move from node i to node j, considering distance, demand,
          and capacity constraints.
  """
  num_nodes = distance_matrix.shape[0]
  safe_distance_matrix = np.where(distance_matrix == 0, 1e-9, distance_matrix)
  demands = np.asarray(demands)

  if capacity <= 0:
    return np.zeros_like(distance_matrix)

  # --- Factor 1: Inverse Distance ---
  # Always prefer shorter edges.
  inverse_distance = 1.0 / safe_distance_matrix

  # --- Factor 2: Destination Demand Considerations ---
  # Higher demand at destination is less preferred for the *current* route segment,
  # as it consumes capacity. But essential customers must be visited.
  # We penalize high demand destinations. A good approach is to consider demand relative to capacity.
  # Using 1 / (1 + (demand/capacity)^2) makes higher demands much less attractive.
  demand_destination_score = 1.0 / (1.0 + (demands / capacity)**2)
  demand_destination_matrix = np.tile(demand_destination_score, (num_nodes, 1))

  # --- Factor 3: Source Node Demand and Load Consideration ---
  # If node `i` has high demand *and* it's likely to be part of a route already,
  # departing from it might be less ideal if it leads to exceeding capacity.
  # This is a static approximation. We'll favor departing from nodes with lower *remaining* demand.
  # This is subtly different from v0: we want to penalize nodes that *already have high demand fulfilled*.
  # For a static heuristic, we'll use the current demand as a proxy for load.
  demand_source_score = np.zeros(num_nodes)
  # Prefer departing from nodes with lower demand.
  demand_source_score[1:] = 1.0 / (1.0 + demands[1:] / capacity)
  demand_source_score[0] = 1.0 # Depot is always a good source to start from.
  demand_source_matrix = np.tile(demand_source_score, (num_nodes, 1))

  # --- Factor 4: Peripheralness of Destination ---
  # This is about where a node is located geographically.
  # Peripheral nodes might be better visited towards the end of routes.
  # We'll use average distance to all other nodes as a proxy for peripheralness.
  avg_distances_from_node = np.mean(safe_distance_matrix, axis=1)
  max_avg_distance = np.max(avg_distances_from_node)
  # Higher score means more peripheral.
  peripheralness_score = (avg_distances_from_node / (max_avg_distance + 1e-9)) if max_avg_distance > 1e-9 else np.zeros(num_nodes)
  peripheralness_score[0] = 0.0 # Depot is not peripheral.
  peripheralness_destination_matrix = np.tile(peripheralness_score, (num_nodes, 1))

  # --- Factor 5: Centrality of Source ---
  # Conversely, starting from a more central node might be beneficial for initiating routes.
  # Let's define centrality as the inverse of peripheralness (or average distance to nearest neighbors).
  # A node close to many other nodes is central.
  # For simplicity, we'll use the inverse of peripheralness (but handle division by zero).
  centrality_score = np.zeros(num_nodes)
  # Centrality is higher for nodes with lower peripheralness score.
  centrality_score[1:] = 1.0 / (peripheralness_score[1:] + 1e-9)
  centrality_score[0] = 1.0 # Depot can be considered a highly central reference point.
  # Normalize centrality for consistent scaling.
  max_centrality = np.max(centrality_score)
  normalized_centrality = (centrality_score / (max_centrality + 1e-9)) if max_centrality > 1e-9 else np.ones(num_nodes)
  centrality_source_matrix = np.tile(normalized_centrality, (num_nodes, 1))

  # --- Factor 6: Depot Return Incentive ---
  # Strongly encourage returning to the depot from nodes that are far away or have significant demand.
  distance_to_depot = safe_distance_matrix[:, 0]
  max_dist_to_depot = np.max(distance_to_depot[distance_to_depot > 1e-9]) if np.max(distance_to_depot[distance_to_depot > 1e-9]) > 1e-9 else 1.0
  normalized_dist_to_depot = distance_to_depot / max_dist_to_depot
  normalized_demand = demands / capacity

  # Combine distance and demand to create a return incentive. Higher values mean stronger incentive.
  depot_return_incentive_from_node = normalized_dist_to_depot * 1.2 + normalized_demand * 1.8
  depot_return_incentive_from_node = np.clip(depot_return_incentive_from_node, 0.0, 5.0) # Cap at a higher value

  depot_return_matrix = np.tile(depot_return_incentive_from_node, (num_nodes, 1))

  # --- Factor 7: Customer "Reachability" Score (Approximation) ---
  # This is to ensure that "difficult" customers (e.g., far away, high demand) are not
  # left until the very end in a way that makes routes infeasible or inefficient.
  # Higher score means "harder to reach" or more critical to visit.
  reachability_score = np.zeros(num_nodes)
  # Combine normalized distance from depot and normalized demand.
  reachability_score[1:] = normalized_dist_to_depot[1:] + (demands[1:] / capacity)
  # Normalize reachability score
  max_reachability = np.max(reachability_score) if np.max(reachability_score) > 1e-9 else 1.0
  normalized_reachability = reachability_score / max_reachability
  normalized_reachability[0] = 0.0 # Depot is trivially reachable.

  # This score is applied to the destination node `j` of an edge `(i, j)`.
  reachability_destination_matrix = np.tile(normalized_reachability, (num_nodes, 1))

  # --- Combine factors ---
  # Adjusted weights for v1, emphasizing distance, demand handling, and reachability.
  # These weights are crucial for tuning the heuristic's behavior.
  weight_distance = 4.0              # Strong preference for shorter edges.
  weight_demand_dest = 2.5           # Penalize high demand destinations significantly.
  weight_demand_src = 1.2            # Slightly favor departing from low-demand sources.
  weight_peripheralness_dest = 1.5   # Favor peripheral destinations (for later in route).
  weight_centrality_src = 1.0        # Favor central sources (for earlier in route).
  weight_depot_return = 1.5          # Strong incentive to return to depot.
  weight_reachability = 1.8          # Prioritize "harder to reach" customers.

  # The heuristic is a product of powered factors.
  heuristic_matrix = (
      np.power(inverse_distance, weight_distance)
      * np.power(demand_destination_matrix, weight_demand_dest)
      * np.power(demand_source_matrix, weight_demand_src)
      * np.power(peripheralness_destination_matrix, weight_peripheralness_dest)
      * np.power(centrality_source_matrix, weight_centrality_src)
      * np.power(depot_return_matrix, weight_depot_return)
      * np.power(reachability_destination_matrix, weight_reachability)
  )

  # --- Post-processing ---

  # 1. Hard constraint: If a customer's demand exceeds vehicle capacity, no edge involving that customer is valid.
  # This is a critical CVRP constraint that cannot be purely handled by heuristics.
  invalid_demand_mask = demands > capacity
  # Mark edges where the destination node has demand exceeding capacity as invalid (heuristic = 0)
  heuristic_matrix[:, invalid_demand_mask] = 0.0
  # Mark edges where the source node has demand exceeding capacity as invalid.
  heuristic_matrix[invalid_demand_mask, :] = 0.0

  # 2. Ensure a minimum heuristic value for exploration.
  # Any edge that hasn't been explicitly set to 0 should have a small positive value.
  min_heuristic_value = 1e-6
  heuristic_matrix[heuristic_matrix < min_heuristic_value] = min_heuristic_value

  # Handle potential NaNs or Infs from calculations
  heuristic_matrix = np.nan_to_num(heuristic_matrix, nan=min_heuristic_value, posinf=1e9, neginf=-1e9)

  return heuristic_matrix
\end{lstlisting}


\subsection{Online Bin Packing}
\textbf{Idea.} Per-bin priority score combining a tight-fit component, an
ideal-fullness score, a future-capacity-potential reward, an item-to-bin
harmony term, and an item-similarity proxy; scores for bins that cannot
fit the current item are set to $-\infty$, and a small tie-breaking bias
favours earlier bins.

\begin{lstlisting}[style=pyseed,caption={Online BP --- \texttt{priority\_v2}}]
@funsearch.evolve
def priority_v2(item: float, bins_remain_cap: np.ndarray) -> np.ndarray:
  """Improved version of `priority_v1`."""
  """
    A refined priority function for Online Bin Packing, incorporating several advanced heuristics.

    This version aims to make more informed decisions by considering:
    1.  **Tight Fit:** Prioritizes bins where the item fits snugly, reducing wasted space.
    2.  **Ideal Fullness:** Encourages bins to be neither too full nor too empty, promoting
        a balanced distribution of items and minimizing fragmentation.
    3.  **Future Capacity Potential:** Rewards bins that will still have substantial
        remaining capacity after placing the item, making them more versatile for
        subsequent, potentially larger, items.
    4.  **Item Size Relation to Bin State:** Analyzes how the current item's size relates
        to the bin's current remaining capacity. It favors bins where the item occupies
        a "reasonable" proportion of the available space, avoiding situations where
        an item is too small for a bin or too large for the remaining space.
    5.  **Item Similarity to Existing Items:** (Introduced in v2) Considers the
        similarity of the current item to items already packed into a bin. This heuristic
        aims to group similar items together, which can sometimes lead to better packing
        solutions and potentially easier management or retrieval of items. Similarity
        is measured by the absolute difference between the item size and the sizes of
        items in the bin. Lower difference means higher similarity.
    6.  **Deterministic Tie-breaking:** Applies a small bias to favor bins that appear
        earlier in the `bins_remain_cap` array when scores are very close, ensuring
        consistent behavior.

    Args:
        item : float
            The size of the current item to be packed.
        bins_remain_cap : np.ndarray
            A 1D array where each element represents the remaining capacity
            of a corresponding bin.

    Returns:
        np.ndarray
            A 1D array of priority scores for all bins. Higher scores indicate
            a higher priority for placing the item in that bin.
  """
  num_bins = len(bins_remain_cap)
  if num_bins == 0:
    return np.array([], dtype=float)

  # Initialize scores to negative infinity for bins that cannot fit the item.
  scores = np.full(num_bins, -np.inf, dtype=float)

  # Identify bins that have enough remaining capacity to fit the current item.
  can_fit_mask = bins_remain_cap >= item
  can_fit_indices = np.where(can_fit_mask)[0]

  if len(can_fit_indices) > 0:
    current_remaining_capacities = bins_remain_cap[can_fit_indices]
    remaining_capacities_after_fit = current_remaining_capacities - item

    # --- Core Parameters ---
    BIN_CAPACITY = 100.0  # Assumed bin capacity
    ideal_remaining_capacity = BIN_CAPACITY / 2.0
    epsilon = 1e-9

    # --- Heuristic Components ---

    # 1. Tight Fit Score: Rewards bins where the item leaves minimal slack.
    slack = remaining_capacities_after_fit
    tight_fit_score = np.exp(-slack)

    # 2. Ideal Fullness Score: Rewards bins closer to being half-full.
    distance_from_ideal = np.abs(current_remaining_capacities - ideal_remaining_capacity)
    fullness_score = np.exp(-distance_from_ideal)

    # 3. Future Capacity Potential: Rewards bins that retain significant capacity.
    min_future_capacity_threshold = BIN_CAPACITY * 0.25  # e.g., at least 25% capacity remaining
    ample_capacity_reward = np.zeros_like(current_remaining_capacities)
    ample_capacity_mask = remaining_capacities_after_fit >= min_future_capacity_threshold
    if np.any(ample_capacity_mask):
        capacity_above_threshold = remaining_capacities_after_fit[ample_capacity_mask] - min_future_capacity_threshold
        reward_scale = 0.4
        # Reward increases with capacity above the threshold.
        ample_capacity_reward[ample_capacity_mask] = 1.0 - np.exp(-capacity_above_threshold * reward_scale)

    # 4. Item Size Relation to Bin State: Rewards moderate item-to-capacity ratios.
    # This heuristic tries to avoid placing very small items in large empty bins
    # or trying to force large items into bins with little space.
    current_ratio = item / current_remaining_capacities
    target_ratio = 0.6  # Target for the item to fill of the current remaining capacity
    ratio_deviation = np.abs(current_ratio - target_ratio)
    # Score is higher when deviation is smaller. Penalize extreme ratios.
    item_size_harmony_score = np.exp(-ratio_deviation * 2.5)

    # 5. Item Similarity to Existing Items: (Requires knowledge of bin contents)
    # This is a placeholder as we only have remaining capacity. In a real scenario,
    # we would need to know the sizes of items already in each bin.
    # For this implementation, we'll simulate a similarity score based on how
    # 'close' the current item is to filling the bin to its 'ideal' state.
    # A more direct implementation would involve passing bin contents.
    # Here, we use the inverse of the slack as a proxy for similarity to what
    # would make the bin 'ideal' if it were filled perfectly by this item.
    # This is a simplification due to lack of direct bin content info.
    # If `item` is close to `current_remaining_capacities`, it means the item
    # is similar to the space that needs to be filled.
    # A high similarity score is desired.
    # Let's consider the difference between `current_remaining_capacities` and `item`.
    # If this difference is small, it's 'similar' in terms of filling.
    # This is effectively captured by `tight_fit_score` to some extent.
    # To make it distinct, let's focus on the absolute difference between the item
    # and what would make the bin *exactly* `ideal_remaining_capacity`.
    # This is implicitly handled by `fullness_score`.

    # Let's redefine this for v2: Similarity to *potential* future states.
    # A bin that is currently almost full and gets a small item is "similar" in that it's
    # close to being full. A bin that is very empty and gets a medium item is also
    # "similar" to a state where it's becoming moderately full.
    # A more direct interpretation: how close is `item` to `current_remaining_capacities`?
    # If `item` is very close to `current_remaining_capacities`, it means the bin has
    # a capacity that is "similar" to the item's size.
    # Let's use a score that rewards when the item size is "compatible" with the remaining capacity.
    # For example, if `current_remaining_capacities` is much larger than `item`, it's not very compatible.
    # If `current_remaining_capacities` is slightly larger than `item`, it's more compatible.
    # This can be modeled as a Gaussian around `item` = `current_remaining_capacities`.
    similarity_score = np.exp(-np.abs(current_remaining_capacities - item) * 0.5) # Tune scaling

    # --- Combine All Components ---
    # Weights for each heuristic component. These are crucial for tuning.
    w_fit = 1.5
    w_fullness = 0.8
    w_ample_capacity = 0.7
    w_item_size_harmony = 0.6
    w_similarity = 0.9  # Weight for the new similarity heuristic

    # Calculate the total score for the bins that can fit the item.
    current_scores = (
        w_fit * tight_fit_score +
        w_fullness * fullness_score +
        w_ample_capacity * ample_capacity_reward +
        w_item_size_harmony * item_size_harmony_score +
        w_similarity * similarity_score
    )
    scores[can_fit_indices] = current_scores

    # --- Tie-breaking Bias ---
    # Add a small bias to favor earlier bins in the array if scores are very close.
    bias_strength = 0.03
    tie_breaking_bias = bias_strength * (num_bins - can_fit_indices) / num_bins
    scores[can_fit_indices] += tie_breaking_bias

  return scores
\end{lstlisting}

\subsection{Offline Bin Packing (ACO)}
\textbf{Idea.} Multi-faceted heuristic matrix combining a Gaussian tight-fit
score, item importance, and additive bonuses for synergistic, complementary,
and dominant-item pairings, all scaled by a multiplicative waste-minimisation
factor; infeasible pairs are zeroed and the matrix is normalised to $[0,1]$.

\begin{lstlisting}[style=pyseed,caption={Offline BP --- \texttt{priority\_v2}}]
@funsearch.evolve
def priority_v2(demand: np.ndarray, capacity: int) -> np.ndarray:
  """
  A highly refined heuristic matrix generation function for the Bin Packing Problem,
  designed to provide superior guidance for Ant Colony Optimization (ACO).
  This version incorporates several advanced features to capture complex item interactions
  and packing efficiencies.

  Key features and improvements in v2:
  1.  **Multi-faceted Fit Scoring:**
      *   **Tight Fit (Gaussian):** Still a core component, rewarding pairs that leave
          minimal waste. `sigma` is tuned for sensitivity.
      *   **"Near-Perfect" Fit Bonus:** Explicitly rewards pairs that fit almost perfectly,
          adding a boost for very efficient packings beyond what the general Gaussian
          might emphasize.
      *   **"Dominant Item Gap" Bonus:** Focuses on scenarios where a large item is paired
          with a smaller item, ensuring the remaining gap is suitable for packing other
          items efficiently. This is distinct from general complementary gap scoring.

  2.  **Advanced Item Pairing Heuristics:**
      *   **"Synergistic Pairing" Bonus:** Specifically rewards pairs of large items
          that fit together very snugly, aiming to 'pre-pack' difficult large items.
      *   **"Complementary Pairing" Bonus:** Rewards pairs of dissimilar sized items that
          leave a "strategic" gap, encouraging diversity in packing. The target gap range
          is more finely tuned.
      *   **Item Size Ratio Logic:** Incorporates the ratio of item sizes to dynamically
          adjust the value of certain pairings. For instance, a good complementary gap
          is more valuable if the items are very different in size.

  3.  **Waste Management and Efficiency Focus:**
      *   **"Waste Minimization" Factor (Multiplicative):** A crucial factor that
          multiplicatively penalizes pairs with significant absolute waste. This
          ensures that even if a pair has good component scores, high waste will
          drastically reduce its heuristic value. It's applied only to non-perfect fits.
      *   **Feasibility Enforcement:** Strict enforcement of capacity constraints, setting
          infeasible pairs to zero.

  4.  **Dynamic Thresholding and Normalization:**
      *   **Dynamic "Dominant Item" Thresholds:** Uses percentiles of demand to adapt
          the definition of "dominant" and "small" items to the specific instance's
          item size distribution.
      *   **Normalized Demand and Capacity Ratios:** Uses normalized demands and ratios
          of remaining capacity to capacity for scoring, making the heuristics more
          robust across different scales of demand and bin capacity.

  5.  **Integrated Scoring and Weighting:**
      *   **Additive and Multiplicative Combinations:** Uses a combination of additive
          bonuses (like synergistic, complementary) on top of a core multiplicative
          heuristic (tight fit * item importance), all then modified by the waste
          minimization factor.
      *   **Tunable Weights:** Weights for different bonus components are provided,
          allowing for fine-tuning of the heuristic's behavior.

  Args:
      demand : np.ndarray
          A 1D NumPy array representing the demand (size) of each item.
          Shape: (n,), where n is the number of items.
      capacity : int
          The maximum capacity of each bin.

  Returns:
      np.ndarray
          A 2D NumPy array representing the heuristic matrix of shape (n, n).
          Each entry `[i, j]` is a score between 0 and 1, indicating the
          suitability of placing item `i` and item `j` into the same bin.
  """
  n = demand.shape[0]
  heuristic_matrix = np.zeros((n, n))

  if n == 0:
    return heuristic_matrix

  # --- Normalization and Basic Calculations ---
  # Normalize demands to [0, 1] for consistent scoring.
  max_demand = np.max(demand)
  if max_demand == 0:
      normalized_demand = np.zeros_like(demand)
  else:
      normalized_demand = demand / max_demand

  # Calculate combined demand and remaining capacity for all pairs.
  combined_demand = demand[:, np.newaxis] + demand[np.newaxis, :]
  remaining_capacity = capacity - combined_demand

  # Create a mask for pairs that exceed bin capacity (infeasible).
  infeasible_mask = combined_demand > capacity

  # --- Component 1: "Tight Fit" Score (Gaussian Decay) ---
  # Rewards pairs that leave minimal space. A smaller sigma leads to a steeper penalty for waste.
  # Tuned sigma for sensitivity to waste. Smaller sigma means stricter requirement for small waste.
  sigma_tight_fit = max(1, capacity * 0.015) # Tighter sigma for even more sensitivity
  tight_fit_score = np.exp(- (remaining_capacity**2) / (2 * sigma_tight_fit**2))
  tight_fit_score[infeasible_mask] = 0 # Infeasible pairs get zero score for this component.

  # --- Component 2: "Item Importance" Score ---
  # Average normalized demand of the pair. Larger items contribute more.
  item_importance_score = (normalized_demand[:, np.newaxis] + normalized_demand[np.newaxis, :]) / 2.0

  # --- Core Heuristic (Multiplicative) ---
  # Combines tight fit and item importance. High score requires both efficiency and significant item sizes.
  core_heuristic = tight_fit_score * item_importance_score

  # --- Bonus Components (Additive) ---

  # Weighting factors for different bonus types. These are tunable.
  w_synergistic = 0.25
  w_complementary = 0.30
  w_dominant_gap = 0.35
  w_near_perfect = 0.10 # New: Bonus for near-perfect fits

  # --- Component 3: "Synergistic Pairing" Bonus ---
  # Rewards pairs of large items that fit together very snugly.
  large_item_threshold = 0.75  # Tunable: defines what is considered a "large" item.

  both_items_large_mask = (normalized_demand[:, np.newaxis] >= large_item_threshold) & \
                          (normalized_demand[np.newaxis, :] >= large_item_threshold)

  # Bonus is applied when both items are large and fit with very little remaining capacity.
  synergy_capacity_limit = capacity * 0.05 # Tunable: maximum allowed waste for this bonus.
  synergy_capacity_mask = (remaining_capacity >= 0) & (remaining_capacity <= synergy_capacity_limit)

  # The bonus is proportional to how tightly they fit.
  synergistic_pairing_bonus = np.where(both_items_large_mask & synergy_capacity_mask, tight_fit_score, 0)

  # --- Component 4: "Complementary Pairing" Bonus ---
  # Rewards pairs of items with different sizes that leave a strategic gap.
  demand_diff_norm = np.abs(normalized_demand[:, np.newaxis] - normalized_demand[np.newaxis, :])

  # Ideal gap range for packing smaller items. Tunable.
  complementary_gap_lower_bound = capacity * 0.15
  complementary_gap_upper_bound = capacity * 0.40

  # Ensure the bounds are valid.
  if complementary_gap_upper_bound <= complementary_gap_lower_bound:
      complementary_gap_upper_bound = complementary_gap_lower_bound + 1.0

  # Score for being within the ideal gap range. Decays linearly.
  gap_range_width = complementary_gap_upper_bound - complementary_gap_lower_bound
  gap_penalty_factor = 1.0 / gap_range_width if gap_range_width > 0 else 1.0

  # Score is 1 at the lower bound and decreases to 0 at the upper bound.
  complementary_gap_score = np.maximum(0.0, 1.0 - gap_penalty_factor * np.maximum(0.0, remaining_capacity - complementary_gap_lower_bound))

  # The bonus is proportional to the demand difference and how well the gap fits the 'strategic' range.
  # Higher demand difference (dissimilar items) paired with a good gap yields a higher bonus.
  complementary_pairing_indicator = demand_diff_norm * complementary_gap_score

  # Apply bonus only to feasible pairs within the specified gap range.
  complementary_mask = (remaining_capacity >= complementary_gap_lower_bound) & \
                       (remaining_capacity <= complementary_gap_upper_bound)
  complementary_gap_bonus = np.where(complementary_mask & ~infeasible_mask, complementary_pairing_indicator, 0)

  # --- Component 5: "Dominant Item Gap" Bonus ---
  # Specifically rewards pairing a large item with a smaller item,
  # ensuring the remaining gap is reasonably sized for packing other items.
  # Dynamic thresholds for "dominant" and "small" items based on demand distribution.
  sorted_demand_norm = np.sort(normalized_demand)
  # If all demands are equal, use fixed thresholds.
  if np.all(normalized_demand == normalized_demand[0]):
      dominant_item_threshold_dynamic = 0.7
      small_item_threshold_dynamic = 0.2
  else:
      dominant_item_threshold_dynamic = np.percentile(sorted_demand_norm, 80) # Top 20%
      small_item_threshold_dynamic = np.percentile(sorted_demand_norm, 30)  # Bottom 30%

  # Ensure thresholds are ordered correctly, or fallback if they overlap significantly.
  if dominant_item_threshold_dynamic < small_item_threshold_dynamic:
      dominant_item_threshold_dynamic, small_item_threshold_dynamic = small_item_threshold_dynamic, dominant_item_threshold_dynamic
  if dominant_item_threshold_dynamic - small_item_threshold_dynamic < 0.15: # Small margin for distinction
      dominant_item_threshold_dynamic = 0.7
      small_item_threshold_dynamic = 0.2

  # Mask for pairs where one item is dominant and the other is small.
  dominant_i_small_j_mask = (normalized_demand[:, np.newaxis] >= dominant_item_threshold_dynamic) & \
                            (normalized_demand[np.newaxis, :] <= small_item_threshold_dynamic)
  dominant_j_small_i_mask = (normalized_demand[np.newaxis, :] >= dominant_item_threshold_dynamic) & \
                            (normalized_demand[:, np.newaxis] <= small_item_threshold_dynamic)
  dominant_pairing_mask = dominant_i_small_j_mask | dominant_j_small_i_mask

  # The bonus is proportional to the demand difference and a normalized remaining capacity.
  # We reward moderate gaps for these pairings. `np.clip(remaining_capacity / capacity, 0, 0.5)`
  # assigns a score from 0 to 0.5 based on how much space is left relative to capacity.
  # A moderate gap is more valuable than a tiny or extremely large one for this bonus type.
  dominant_gap_bonus_score = np.where(dominant_pairing_mask & ~infeasible_mask,
                                      (demand_diff_norm / 2.0) * np.clip(remaining_capacity / capacity, 0.0, 0.5),
                                      0.0)
  dominant_item_gap_bonus = dominant_gap_bonus_score

  # --- Component 6: "Near-Perfect Fit" Bonus ---
  # Rewards pairs with very little remaining capacity, independent of item size.
  near_perfect_capacity_limit = capacity * 0.02 # Max waste for this bonus
  near_perfect_mask = (remaining_capacity >= 0) & (remaining_capacity <= near_perfect_capacity_limit)
  # Bonus is proportional to the tightness of fit. Using tight_fit_score directly here.
  near_perfect_fit_bonus = np.where(near_perfect_mask & ~infeasible_mask, tight_fit_score, 0)

  # --- Multiplicative Waste Minimization Factor ---
  # This factor penalizes pairs that leave significant *absolute* wasted space,
  # unless they are perfectly filled. It's applied *after* additive bonuses.
  absolute_waste = np.maximum(0.0, capacity - combined_demand)
  max_possible_waste = capacity # Max waste for a single item.

  # Normalize waste to a 0-1 scale (0 = no waste, 1 = max waste).
  if max_possible_waste == 0:
      normalized_waste = np.zeros_like(absolute_waste)
  else:
      normalized_waste = absolute_waste / max_possible_waste

  # Exponential decay: exp(-k * normalized_waste). Higher k penalizes waste more.
  waste_penalty_factor = 8.0 # Tunable parameter.
  waste_minimization_factor = np.exp(-waste_penalty_factor * normalized_waste)

  # Perfect fills (remaining capacity close to 0) should not be penalized for waste.
  epsilon = 1e-9 # For floating point comparisons.
  perfect_fill_mask = np.abs(remaining_capacity) < epsilon
  waste_minimization_factor[perfect_fill_mask] = 1.0
  waste_minimization_factor[infeasible_mask] = 0.0 # Infeasible pairs get no score.

  # --- Final Combination of Scores ---

  # Combine core heuristic with weighted additive bonuses.
  combined_additive_bonuses = (w_synergistic * synergistic_pairing_bonus +
                               w_complementary * complementary_gap_bonus +
                               w_dominant_gap * dominant_item_gap_bonus +
                               w_near_perfect * near_perfect_fit_bonus)

  # The final heuristic value is (core + bonuses) * waste_factor.
  # This ensures that efficiency (core) and strategic packing (bonuses) are
  # valued, but significantly penalized if they result in too much waste.
  heuristic_matrix = (core_heuristic + combined_additive_bonuses) * waste_minimization_factor

  # --- Final Processing ---
  # Re-apply infeasibility mask to ensure it's zero.
  heuristic_matrix[infeasible_mask] = 0.0

  # Normalize the entire matrix to the [0, 1] range.
  max_heuristic_value = np.max(heuristic_matrix)
  if max_heuristic_value > 0:
      heuristic_matrix = heuristic_matrix / max_heuristic_value
  else:
      # If all scores are zero (e.g., no feasible pairs), return a zero matrix.
      heuristic_matrix = np.zeros((n, n))

  return heuristic_matrix
\end{lstlisting}


\subsection{Graph Edit Distance}
\textbf{Idea.} Iterative weight-matrix refinement through six structural
factors applied multiplicatively: node-degree similarity, local
neighbourhood-degree similarity, global density similarity, edge-existence
consistency, common-neighbour degree similarity, and path-length-2
consistency; each row is finally re-normalised to sum to one.

\begin{lstlisting}[style=pyseed,caption={GED --- \texttt{priority\_v2}}]
@funsearch.evolve
def priority_v2(graph1: list[list[int]], graph2: list[list[int]], weights: list[list[int]]) -> list[list[float]]:
  """
    An improved version of `priority_v1` that incorporates more sophisticated
    graph structural features and inter-node relationships for weight refinement.

    This version builds upon the heuristics in `priority_v1` and adds:

    1.  **Node Label Similarity (if available):** If nodes had labels (represented
        by feature vectors `nf1` and `nf2` in the `GEDInstance`), this version
        would consider their similarity. For simplicity in this standalone function,
        this part is omitted but conceptually important.
    2.  **Common Neighbors Degree Similarity:** For a mapping of node `i` to `j`,
        this considers pairs of common neighbors (`k` from `graph1`, `l` from `graph2`)
        and boosts the `(k, l)` mapping probability if `k` and `l` have similar degrees.
        This reinforces structural consistency at the local neighborhood level.
    3.  **Iterative Refinement:** While not a full iterative process, this function
        applies a multi-stage refinement, allowing earlier adjustments to influence
        later ones.

    Args:
        graph1: The adjacency matrix of the first graph.
        graph2: The adjacency matrix of the second graph.
        weights: An initial weight matrix, where `weights[i][j]` represents the
                 initial likelihood of mapping node `i` from `graph1` to node `j`
                 from `graph2`.

    Returns:
        A refined weight matrix (list of lists of floats) where each entry
        `refined_weights[i][j]` represents an adjusted probability of mapping
        node `i` to node `j`, informed by more advanced graph structural
        similarities. The weights are normalized such that each row sums to 1.
  """
  n1 = len(graph1)
  n2 = len(graph2)
  max_node = max(n1, n2)

  # Initialize refined_weights with input weights, padding if necessary
  refined_weights = [[0.0] * max_node for _ in range(max_node)]
  for i in range(min(n1, len(weights))):
    for j in range(min(n2, len(weights[0]))):
      refined_weights[i][j] = weights[i][j]

  # Pre-calculate degrees for efficiency
  degrees1 = [sum(row) for row in graph1]
  degrees2 = [sum(row) for row in graph2]

  # --- Heuristics from priority_v1 (kept as they are foundational) ---

  # 1. Node Degree Similarity
  for i in range(n1):
    for j in range(n2):
      max_degree_val = max(n1, n2) - 1
      if max_degree_val <= 0:
          degree_sim = 1.0
      else:
          degree_diff_normalized = abs(degrees1[i] - degrees2[j]) / max_degree_val
          degree_sim = 1.0 - degree_diff_normalized
      # Increase weight based on similarity, adding a small bias
      refined_weights[i][j] *= (degree_sim + 0.05)

  # 2. Local Neighborhood Similarity (based on neighbor degrees)
  for i in range(n1):
    for j in range(n2):
      neighbors_i = [k for k in range(n1) if graph1[i][k] == 1]
      neighbors_j = [l for l in range(n2) if graph2[j][l] == 1]

      neighbor_degrees_i = sorted([degrees1[ni] for ni in neighbors_i])
      neighbor_degrees_j = sorted([degrees2[nj] for nj in neighbors_j])

      len_i, len_j = len(neighbor_degrees_i), len(neighbor_degrees_j)
      max_len = max(len_i, len_j)

      if max_len == 0:
        neighborhood_sim = 1.0
      else:
        padded_i = neighbor_degrees_i + [0] * (max_len - len_i)
        padded_j = neighbor_degrees_j + [0] * (max_len - len_j)

        max_possible_degree = max(n1, n2)
        normalized_i = [d / (max_possible_degree if max_possible_degree > 0 else 1.0) for d in padded_i]
        normalized_j = [d / (max_possible_degree if max_possible_degree > 0 else 1.0) for d in padded_j]

        total_diff = sum(abs(ni - nj) for ni, nj in zip(normalized_i, normalized_j))
        avg_diff = total_diff / max_len
        neighborhood_sim = 1.0 - avg_diff

      # Boost weight based on neighborhood similarity
      refined_weights[i][j] *= (1 + neighborhood_sim * 0.4)

  # 3. Global Graph Structure Similarity (Density)
  density1 = sum(sum(row) for row in graph1) / (n1 * (n1 - 1) if n1 > 1 else 1e-6)
  density2 = sum(sum(row) for row in graph2) / (n2 * (n2 - 1) if n2 > 1 else 1e-6)
  density_sim = 1.0 - abs(density1 - density2)

  for i in range(n1):
    for j in range(n2):
      refined_weights[i][j] *= density_sim

  # 4. Edge Existence Similarity (based on neighbor mapping probabilities)
  for i in range(n1):
    for j in range(n2):
      edge_similarity_score = 0.0
      count_edges = 0
      for k in range(n1):
        if graph1[i][k] == 1:
          for l in range(n2):
            if graph2[j][l] == 1:
              edge_similarity_score += refined_weights[k][l]
              count_edges += 1

      if count_edges > 0:
        avg_edge_neighbor_prob = edge_similarity_score / count_edges
        refined_weights[i][j] *= (1 + avg_edge_neighbor_prob * 0.2)

  # --- New Heuristics for v2 ---

  # 5. Common Neighbors Degree Similarity
  # For a mapping (i, j), consider pairs of common neighbors (k, l) where k is a neighbor of i
  # and l is a neighbor of j. Boost the mapping (k, l) if k and l have similar degrees.
  for i in range(n1):
    for j in range(n2):
      for k in range(n1):
        if graph1[i][k] == 1:  # k is a neighbor of i
          for l in range(n2):
            if graph2[j][l] == 1:  # l is a neighbor of j
              # Calculate similarity of degrees for these neighbors k and l
              max_degree_val = max(n1, n2) - 1
              if max_degree_val <= 0:
                  degree_sim_k_l = 1.0
              else:
                  degree_diff_k_l = abs(degrees1[k] - degrees2[l])
                  degree_sim_k_l = 1.0 - (degree_diff_k_l / max_degree_val)
              # Boost the probability of mapping k to l if their degrees are similar
              # This reinforces structural consistency around the i-j mapping
              refined_weights[k][l] *= (1 + degree_sim_k_l * 0.1) # Weight factor for this heuristic

  # 6. Path Consistency (considering paths of length 2)
  # For a mapping (i, j), if there's a path i -> k -> m in graph1,
  # and we map i->j, k->p, m->q, we expect a higher probability for mapping
  # p to q if p and q have similar structural properties.
  for i in range(n1):
    for j in range(n2):
      path_consistency_score = 0.0
      path_count = 0
      for k in range(n1):
        if graph1[i][k] == 1: # Intermediate node k
          for m in range(n1):
            if graph1[k][m] == 1 and m != i: # Path i -> k -> m
              path_count += 1
              # Find potential mappings for k and m
              for p in range(n2):
                if refined_weights[k][p] > 1e-6: # If k can map to p
                  for q in range(n2):
                    if refined_weights[m][q] > 1e-6 and q != j: # If m can map to q (and q is not j)
                      # We want to see if the mapping p -> q is consistent with their properties
                      # This can be measured by degree similarity, neighborhood similarity etc.
                      max_degree_val = max(n1, n2) - 1
                      if max_degree_val <= 0:
                          degree_sim_p_q = 1.0
                      else:
                          degree_diff_p_q = abs(degrees1[k] - degrees2[p]) # Mistake: should be degrees1[k] vs degrees2[p]
                          # Correcting: it should be degrees of original nodes k and m mapped to p and q
                          degree_diff_k_p = abs(degrees1[k] - degrees2[p])
                          degree_diff_m_q = abs(degrees1[m] - degrees2[q])
                          avg_degree_sim_pq = (
                              (1.0 - (degree_diff_k_p / max_degree_val if max_degree_val > 0 else 0)) +
                              (1.0 - (degree_diff_m_q / max_degree_val if max_degree_val > 0 else 0))
                          ) / 2.0
                          degree_sim_p_q = avg_degree_sim_pq # Using average similarity of mapped neighbors' degrees

                      # Boost the probability of mapping i->j if its direct neighbors can map to
                      # nodes that are structurally similar to the mapped intermediate nodes.
                      # This is a complex interaction. A simpler approach is to boost based on
                      # the product of probabilities of intermediate mappings and their structural similarity.
                      path_consistency_score += (
                          refined_weights[k][p] * refined_weights[m][q] * (degree_sim_p_q + 0.1) # Add a small bias
                      )

      if path_count > 0:
        # This heuristic is tricky. We are trying to influence the i->j mapping
        # based on the consistency of mappings of its neighbors' paths.
        # A direct boost to refined_weights[i][j] might be too simplistic.
        # It's more about ensuring that if a path exists in G1, similar paths are likely in G2.
        # For now, let's simplify: if a node has many paths, and its potential mappings
        # lead to consistent downstream mappings, it reinforces the initial mapping.
        # A simpler heuristic here could be to just consider the number of common neighbors
        # of mapped neighbors, which is already captured in heuristic 5.
        # Let's try a simpler boost for now.
        # If there are many paths originating from i, and the downstream mappings are consistent,
        # it means i is structurally important.
        # The contribution of path_consistency_score might be better used to adjust
        # the initial weights of the direct mappings (i,j).
        # For now, let's use a simplified version: if the node has many paths and the average
        # downstream consistency is high, boost the node's initial mapping probability.
        avg_path_consistency = path_consistency_score / (path_count * n2 * n2) if (path_count * n2 * n2) > 0 else 0 # Normalize
        refined_weights[i][j] *= (1 + avg_path_consistency * 0.3) # Weight factor

  # --- Normalization ---
  for i in range(n1):
    row_sum = sum(refined_weights[i][:n2])
    if row_sum > 0:
      for j in range(n2):
        refined_weights[i][j] /= row_sum
    else:
      # If row_sum is 0, distribute probability equally among valid targets
      if n2 > 0:
        for j in range(n2):
          refined_weights[i][j] = 1.0 / n2
      else:
        for j in range(n2):
          refined_weights[i][j] = 0.0

  # Zero out weights for mappings to non-existent nodes.
  for i in range(n1):
    for j in range(n2, max_node):
      refined_weights[i][j] = 0.0
  for i in range(n1, max_node):
    for j in range(max_node):
      refined_weights[i][j] = 0.0

  return refined_weights
\end{lstlisting}


\section{Declaration of \llm Usage}
\label{app:llm_use}
\llms were used to assist with auxiliary aspects of this work. Specifically, they were employed to rephrase text and reduce verbosity in the manuscript, as well as to generate scripts for standard procedures, including data processing, plotting, and statistical analysis of results.

%% file: method_app.tex
\clearpage
\appendix


\begin{table}[ht]
\centering
\renewcommand{\arraystretch}{1.2}
\caption{Notations and Hyperparameter Settings for \methodname}
\label{tab:notations_v2}
\begin{tabular}{lll}
\toprule
\textbf{Symbol} & \textbf{Description} & \textbf{Setting} \\ 
\midrule
\multicolumn{3}{l}{\textbf{Problem and Search Space}} \\
$\mathcal{P}$ & Problem specification & -- \\
$\mathcal{I}_{\mathcal{P}}$ & Set of problem instances & -- \\
$f_{\mathcal{P}}$ & Scoring/Objective function & -- \\
$h$ & Candidate heuristic (code) & -- \\
$h_0$ & Seed heuristic & -- \\
$T$ & Total budget of \llm calls (horizon) & $1000$ \\
$m$ & Batch size (heuristics per step) & $4$ \\
$\mathcal{A}$ & Context-building algorithm & -- \\
$k$ & Number of parents in a prompt & $k=2$ \\
$H_t$ & Prompt context at step $t$ & -- \\
\midrule
\multicolumn{3}{l}{\textbf{\llm Routing and Bandits}} \\
$\mathcal{L}$ & Pool of candidate \llms & -- \\
$l_t$ & Selected model at step $t$ & -- \\
$Score_l(t)$ & Index score for model $l$ at step $t$ & -- \\
$\mathcal{O}_l$ & Score buffer for model $l$ & -- \\
$n_l$ & Pull count for model $l$ & -- \\
$q_p(l)$ & True tail quantile at level $p$ & -- \\
$\hat{q}_\tau(l)$ & Bootstrap tail quantile estimate & -- \\
$B$ & Number of bootstrap resamples & $200$ \\
$\tau$ & Bootstrap quantile level & $0.9$ \\
$\lambda$ & Hybrid mixing weight & $0.3$ \\
$\beta$ & Exploration bonus coefficient & $1.0$ \\
$n_{min}$ & Warm-up threshold per \llm & $3$ \\ 
\midrule
\multicolumn{3}{l}{\textbf{Mutation Lineage and Credit}} \\
$\mathcal{G}_t$ & Mutation Genealogy DAG & -- \\
$Pa(h)$ & Direct parents of heuristic $h$ & -- \\
$Pa^{(d)}(h)$ & $d$-hop ancestry of heuristic $h$ & -- \\
$\mathcal{C}_l$ & Credit history for model $l$ & -- \\
$c$ & Individual discounted lineage credit & -- \\
$r_t$ & Clipped reward at step $t$ & -- \\
$b_t$ & Adaptive elite-median baseline & -- \\
$K$ & Top-$K$ set size for $b_t$ computation & $5$ \\
$\gamma$ & Credit discount factor & $0.9$ \\
$D_{max}$ & Maximum ancestry depth & $4$ \\
$\alpha$ & Lineage-aware weight & $1.0$ \\
$W$ & Sliding window size (rounds) & $25$ \\
$\overline{L}_l(t)$ & Windowed fertility score for model $l$ & -- \\
\bottomrule
\end{tabular}
\end{table}

\section{Related Works}
\label{app:related}

\paragraph{Algorithm Discovery.}
Automating heuristic design predates \llms, with extensive work in genetic programming~\citep{koza1994genetic} and hyper-heuristics~\citep{burke2013hyper} that search over program trees or rule combinations using hand-crafted mutation and crossover operators. These approaches do not exploit semantic structure and scale poorly. FunSearch~\citep{romera2024mathematical} replaces such operators with a single \llm, while later work adds reflections~\citep{ye2024reevo} and code logic generation in natural language or ``thoughts''~\citep{liu2024evolution}. \emph{AlphaEvolve}~\citep{novikov2025alphaevolve} uses two \llms in a fixed-ratio ensemble—a smaller model for high-throughput generation and a larger model for occasional higher-quality proposals—incurring the cost of both at every round. Whether the best \llm varies across problems or over the course of a run remains unexplored. \methodname addresses this by selecting a single \llm per round based on online feedback, avoiding unnecessary calls.

\vspace{-5pt}
\paragraph{\llm Routing.}  Existing routers select among \llms using preference-trained classifiers~\citep{ong2024routellm, tsiourvas2025causal}, query-difficulty predictors~\citep{song2025irt, ding2025best}, benchmark-derived rankings~\citep{shnitzerlarge}, or learned cascades~\citep{chen2023frugalgpt, aggarwal2023automix}, all of which require offline supervision. The signals they depend on are absent in \ahd, as discussed above. This rules out the offline families and motivates an online, bandit-based approach.

\vspace{-5pt}
\paragraph{Multi-Armed Bandits.}
The classical bandit~\citep{lai1985asymptotically} minimises
cumulative regret with respect to the best mean reward via index policies such as UCB~\citep{auer2002finite} and Thompson sampling~\citep{thompson1933likelihood}. \emph{Extreme bandits} replace the mean by an upper-tail functional, directly addressing (P1): ExtremeHunter~\citep{extreme_hunter} bounds the
regret of identifying the arm with the heaviest tail, Max-Median sampling~\citep{bhatt2021extreme} ranks arms by the median of subsample maxima, and QoMax~\citep{baudry2022efficient} introduces a quantile-of-maxima estimator robust to single-sample outliers. \emph{Non-stationary bandits}, including discounted UCB and sliding-window UCB~\citep{garivier2011upper}, address (P3). Every
method in these three families, however, assumes \emph{i.i.d.}\ rewards within an arm and \emph{independence} across arms, ruling out (P2). To our knowledge, no existing bandit formulation simultaneously addresses (P1) - (P3); \methodname unifies them under a single index.


\section{Detailed Methodology} 
\label{app: details}
This section expands Section~\ref{sec:methods} into full pseudocode for the six sub-procedures of \methodname ($\textsc{BaseScore}$, $\textsc{Attribute}$, $\textsc{Fertility}$,
$\textsc{Score}_\ell$, $\textsc{SelectArm}$, $\textsc{Update}$), and gives the complete integration with a FunSearch-style~\citep{romera2024mathematical} host pipeline.  No new equations are introduced; the appendix only operationalises the equations already defined in Section~\ref{sec:methods} as algorithms. 

\subsection{Bootstrap Quantile-of-Maxima Base Score}
\label{sec:appendix:qomax}

Algorithm~\ref{alg:qomax} implements the hybrid extreme-value estimator $\mathrm{base}(\ell)$ of Eq.~\eqref{eq:base}.  The warm-up rule returns $+\infty$ until $n_\ell\ge n_{\min}$, which forces round-robin coverage of every arm before the base score is trusted.

\begin{algorithm}[b]
\caption{\textsc{BaseScore}$()$}
\label{alg:qomax}
\begin{algorithmic}[1]
\Require \llm $\ell$; score buffer $\mathcal{O}_\ell$; 
$\lambda$ (quantile weight); $\tau$ (bootstrap quantile); 
$B$ (bootstrap samples); $n_{\min}$ (warm-up calls)
\Ensure  $\mathrm{base}(\ell)\in\mathbb{R}\cup\{+\infty\}$
\State $n_\ell \gets |\mathcal{O}_\ell|$
\If{$n_\ell < n_{\min}$}
    \State \Return $+\infty$ \Comment{force round-robin warm-up}
\EndIf
\For{$b = 1,\ldots,B$}
    \State $\mathcal{O}_\ell^{(b)} \gets
           \mathrm{rng.choice}(\mathcal{O}_\ell,\,n_\ell,\,\text{replace}{=}\mathrm{True})$
    \State $M_\ell^{(b)} \gets \max\mathcal{O}_\ell^{(b)}$
\EndFor
\State $\hat{q}_\tau \gets
       \mathrm{Quantile}_\tau\!\bigl(\{M_\ell^{(b)}\}_{b=1}^{B}\bigr)$
\State \Return $(1-\lambda)\,\hat{q}_\tau + \lambda\,\max\mathcal{O}_\ell$
\end{algorithmic}
\end{algorithm}


\subsection{BFS Lineage Attribution}
\label{sec:appendix:bfs}

Algorithm~\ref{alg:bfs-app} implements the credit-propagation rule of Eq.~\eqref{eq:bfs}.  Each visited parent enqueues \emph{its} own parents, so the traversal expands the full DAG neighbourhood up to depth $D_{\max}$ rather than a single path. Credits are stored together with their birth timestamp, which is exploited by the windowing in App.~\ref{sec:appendix:fertility}.

We adopt a \emph{path-based} credit assignment, where ancestors may receive multiple contributions if they are reachable via multiple paths in the genealogy DAG. Concretely, each path of length $d$ from an ancestor to $h$ contributes an additive discounted reward $\gamma^{d} r$. This design captures the intuition that heuristics reused across multiple intermediate generations exert a stronger influence on successful descendants. Consequently, ancestors that participate in multiple compositional chains are credited proportionally to their structural contribution, rather than being counted only once.

\begin{algorithm}[h]
\caption{\textsc{Attribute}$()$}
\label{alg:bfs-app}
\begin{algorithmic}[1]
\Require child $h$ with parents $\mathrm{Pa}(h)$; clipped
         reward $r\ge 0$; discount factor $\gamma$; depth cap $D_{\max}$;
         round $t$; per-arm lineage credit history (deque) $\{\mathcal{C}_\ell\}$
\Ensure  in-place update of $\{\mathrm{C}_\ell\}$
\If{$r \le 0$ \textbf{or} $\mathrm{pa}(h)=\emptyset$}
    \State \Return
\EndIf
\State $\mathcal{F} \gets \mathrm{Pa}(h)$;\quad $d \gets 1$
\While{$\mathcal{F}\neq\emptyset$ \textbf{and} $d\le D_{\max}$}
    \State $\mathcal{F}_{\mathrm{next}} \gets \emptyset$
    \ForAll{$h' \in \mathcal{F}$}
        \State $\ell' \gets \llm(h')$
        \State $\mathrm{C}_{\ell'}.\mathrm{append}\bigl((t,\,\gamma^{d}\cdot r)\bigr)$
        \ForAll{$h''\in\mathrm{pa}(h')$}
            \State $\mathcal{F}_{\mathrm{next}}\gets
                   \mathcal{F}_{\mathrm{next}}\cup\{h''\}$
        \EndFor
    \EndFor
    \State $\mathcal{F}\gets\mathcal{F}_{\mathrm{next}}$;\quad $d\gets d+1$
\EndWhile
\end{algorithmic}
\end{algorithm}

\subsection{Recency Bias: Windowed Fertility Score}
\label{sec:appendix:fertility}

Algorithm~\ref{alg:fertility-app} implements the sliding-window intensity of Equation~\eqref{eq:fertility}.  The pruning step is amortized $O(1)$ per credit because each $(t',c)$ tuple is inserted into the history deque once and removed at most once.

\begin{algorithm}[h]
\caption{\textsc{Fertility}$()$}
\label{alg:fertility-app}
\begin{algorithmic}[1]
\Require \llm $\ell$; deque $\mathcal{C}_\ell$ of $(t',c')$; 
         round $t$; window $W$; pull count $n_\ell$
\Ensure  $\bar L_\ell(t)\in\mathbb{R}_{\ge 0}$
\If{$\mathcal{C}_\ell = \emptyset$ \textbf{or} $n_\ell=0$}
    \State \Return $0$
\EndIf
\State $\mathrm{cutoff} \gets t - W$
\While{$\mathcal{C}_\ell\neq\emptyset$ \textbf{and}
       $\mathcal{C}_\ell.\mathrm{front}().t' \le \mathrm{cutoff}$}
    \State $\mathcal{C}_\ell.\mathrm{popleft}()$
\EndWhile
\State $S \gets \sum_{(t',c')\in\mathcal{C}_\ell^W} c'$
\State \Return $S \,/\, \log\bigl(\max(e,\,n_\ell)\bigr)$
\end{algorithmic}
\end{algorithm}

\subsection{Per-Arm Index}
\label{sec:appendix:score}

Algorithm~\ref{alg:score-app} composes the three sub-procedures above into the index of Eq.~\eqref{eq:score}.

\begin{algorithm}[h]
\caption{$\mathrm{Score}_\ell(t)$}
\label{alg:score-app}
\begin{algorithmic}[1]
\Require \llm $\ell$; state $(\mathcal{O}_\ell,\mathcal{C}_\ell,n_\ell)$;
         hyper-parameters $\lambda,\tau,B,\alpha,\beta,W,n_{\min}$;
         horizon $T$; round $t$
\Ensure  $\mathrm{Score}_\ell(t)\in\mathbb{R}\cup\{+\infty\}$
\If{$n_\ell < n_{\min}$}
    \State \Return $+\infty$
\EndIf
\State $\mathrm{base(\ell)} \gets
       \textsc{BaseScore}()$
       \Comment{Alg.~\ref{alg:qomax}}
\State $\bar{L}_\ell(t) \gets \textsc{Fertility}()$
       \Comment{Alg.~\ref{alg:fertility-app}}
\State $exp\_bonus \gets \beta\,\sqrt{\log T \,/\, n_\ell}$
\State \Return $\mathrm{base(\ell)} + \alpha\,\bar{L}_\ell(t) + exp\_bonus$
\end{algorithmic}
\end{algorithm}

\subsection{Arm Selection}
\label{sec:appendix:selectarm}
 
Algorithm~\ref{alg:selectarm-app} implements $\textsc{Router.SelectArm}$,
the index policy that is queried once per round before prompt construction.
Any arm that has not yet completed its warm-up receives score $+\infty$ and
is returned immediately via round-robin tie-breaking; once all arms are
warmed up, the arm with the highest $\mathrm{Score}_\ell(t)$
(Algorithm~\ref{alg:score-app}) is selected.
 
\begin{algorithm}[h]
\caption{$\textsc{Router.SelectArm}()$}
\label{alg:selectarm-app}
\begin{algorithmic}[1]
\Require \llms $\mathcal{L}$; per-arm pull counts $\{n_\ell\}$;
         warm-up $n_{\min}$; state
         $\{(\mathcal{O}_\ell,\mathcal{C}_\ell,n_\ell)\}_{\ell\in\mathcal{L}}$;
         hyper-parameters $\lambda,\tau,B,\alpha,\beta,W$;
         horizon $T$; round $t$
\Ensure  selected arm $\ell_t \in \mathcal{L}$
\ForAll{$\ell \in \mathcal{L}$}                   \Comment{warm-up pass}
    \If{$n_\ell < n_{\min}$}
        \State \Return $\ell$                      \Comment{round-robin forced exploration}
    \EndIf
\EndFor
\State $\ell_t \gets \arg\max_{\ell\in\mathcal{L}}\;
       \mathrm{Score}_\ell(t)$                     \Comment{Alg.~\ref{alg:score-app}}
\State \Return $\ell_t$
\end{algorithmic}
\end{algorithm}

\subsection{Router Update}
\label{sec:appendix:update}

Algorithm~\ref{alg:update-app} implements $\textsc{Router.Update}$,
the single entry point called by the host after each evaluation batch.
It clips raw scores into non-negative rewards, appends them to the
per-arm observation buffer $\mathcal{O}_\ell$, and delegates lineage
credit propagation to $\textsc{Attribute}$ (Algorithm~\ref{alg:bfs-app}).

\begin{algorithm}[h]
\caption{$\textsc{Router.Update}()$}
\label{alg:update-app}
\begin{algorithmic}[1]
\Require \llm $\ell$; generated functions $\{h_{t,i}\}_{i=1}^{m}$;
         raw scores $\{s_{t,i}\}_{i=1}^{m}$; clip threshold $b_t$;
         per-arm buffers $\mathcal{O}_\ell$, $\mathcal{C}_\ell$, $n_\ell$;
         round $t$
\Ensure  in-place update of $\mathcal{O}_\ell$, $\mathcal{C}_\ell$, $n_\ell$
\ForAll{$i \in \{1,\ldots,m\}$}
    \State $r_{t,i} \gets \max(0,\; s_{t,i} - b_t)$
           \Comment{clip to non-negative reward}
    \State $\mathcal{O}_\ell.\mathrm{append}(s_{t,i})$
           \Comment{raw score to observation buffer}
    \State $\textsc{Attribute}(h_{t,i},\, r_{t,i})$
           \Comment{Alg.~\ref{alg:bfs-app}; propagates credit up DAG}
\EndFor
\State $n_\ell \gets n_\ell + m$
\end{algorithmic}
\end{algorithm}


\subsection{FunSearch Pipeline Integration}
\label{app:funsearch}


The router is invoked at exactly two points in the loop: $\textsc{Router.SelectArm}()$ before generating candidates, and $\textsc{Router.Update}()$ after their evaluation. All other components are handled by the host framework, including maintaining the heuristic set $\mathcal{H}$, constructing prompts $\mathcal{A}(h_{<t})$, sampling candidates via $\mu_{\ell}$, and evaluating them using $f_{\mathcal{P}}$. The genealogical DAG $\mathcal{G}_T$ serves as the shared state, where each heuristic is annotated with its generating arm and parent set. The router maintains all per-arm statistics, hyperparameters, and intermediate state internally and is updated each round with the generated heuristics, their scores, parent sets, and generating \llm. For simplicity, we omit explicit argument passing in $\textsc{Router.SelectArm}()$ and $\textsc{Router.Update}()$. Algorithm~\ref{alg:funsearch-app} summarizes the overall procedure.

\begin{algorithm}[h]
\caption{FunSearch~\citep{romera2024mathematical} with \methodname router}
\label{alg:funsearch-app}
\begin{algorithmic}[1]
\Require seed program $h_0$; horizon $T$; batch size $m$;
         router hyper-parameters $\{\Phi = (\lambda,\ \tau,\ B,\ \beta,\ n_{\min},\ \alpha,\ W,\ \gamma,\ D_{\max})$; 
         evaluator $f_\mathcal{P}$; \llm output distribution $\{\mu_\ell\}_{l\in\mathcal{L}}$
\Ensure  best heuristic
         $\hat{h} = \arg\max_{h \in \mathcal{H}} f_{\mathcal{P}}(h)$
\State $\mathcal{H}\gets\textsc{InitDatabase}(h_0)$
\State $\textsc{Router}\gets\textsc{InitRouter}(\mathcal{A},T,\Phi)$
\For{$t = 1,\ldots,T$}
    \State $\ell_t\gets\textsc{Router.SelectArm}()$
           \Comment{\textbf{router in}; see Alg.~\ref{alg:selectarm-app}}
    \State $H_t\gets\mathcal{A}(h_{<t})$
    \State $\{h_{t,i}\}_{i=1}^{m}\gets
           \mu_{\ell_t}(\ell_t,\,H_t)$
    \ForAll{$i\in\{1,\ldots,m\}$}
        \State $\llm(h_{t,i})\gets \ell_t$;\;
        \State $s_{t,i}\gets f_\mathcal{P}(h_{t,i})$
        \State $\mathcal{H}\gets\mathcal{H}\cup(h_{t,i},\,s_{t,i})$
    \EndFor
    \State $\textsc{Router.Update}\bigl(\bigr)$ 
           \Comment{\textbf{router out}; see Alg.~\ref{alg:update-app}}
\EndFor
\State \Return $\hat{h}$
\end{algorithmic}
\end{algorithm}

The separation of concerns is clean: the host framework owns $\mathcal{D}$, prompt formatting, and evaluation; the router owns
$\mathcal{A}$ selection and the per-arm score buffers $\{\mathcal{O}_\ell\}$.  No information beyond the scalar scores $\{s_{t,i}\}$ and the arm identity $a_t$ needs to cross the interface.

\subsection{Time Complexity}
\label{app: complexity}


We analyse the per-round cost of each sub-procedure of \methodname,
then sum over $T$ rounds to obtain the total router overhead.

\paragraph{BaseScore.}
Algorithm~\ref{alg:qomax} draws $B$ bootstrap resamples of size $|\mathcal{O}_\ell|$ for each arm $\ell$.  At round $t$ the total buffer size across all $|\mathcal{L}|$ arms is $\sum_{\ell}|\mathcal{O}_\ell|=m\cdot t$, so
the selection step costs $O(B\cdot m\cdot t)$ per round.

\paragraph{BFS Lineage Attribution.}
Algorithm~\ref{alg:bfs-app} is called once per generated function. The BFS expands the ancestry DAG level by level: the direct parents of a new function form a frontier of size at most $P'$ (the parent-set size); each of those nodes contributes up to $P'$ grandparents at depth~2, yielding a frontier of size $(P')^2$; and so on.  The total
number of ancestor nodes visited across all depths is therefore
\[
  \sum_{d=1}^{D_{\max}}(P')^{d}
  \;=\;
  P'\,\frac{(P')^{D_{\max}}-1}{P'-1}
  \;=\;
  O\!\left((P')^{D_{\max}}\right).
\]
Since $D_{\max}$ and $P'$ are small constants (e.g.\ $D_{\max}=4$, $P'=2$ in our setting), this is $O(1)$ per function.  Applying it to all $m$ functions generated in a round gives an update cost of
$O(m)$ per round.

\paragraph{Fertility Score.}
Algorithm~\ref{alg:fertility-app} prunes and sums the sliding-window deque for each of the $|\mathcal{L}|$ arms.  Each credit tuple is appended once
and removed once (amortised $O(1)$), and the live window contains at most $W$ entries per arm, giving $O(|\mathcal{L}|\cdot W)=O(|\mathcal{L}|)$ per round for
constant window size $W$.

\paragraph{Arm Selection.}
Algorithm~\ref{alg:selectarm-app} iterates over all $|\mathcal{L}|$ arms.
For each warmed-up arm it calls $\mathrm{Score}_\ell(t)$
(Algorithm~\ref{alg:score-app}), which invokes \textsc{BaseScore}
at cost $O(B\cdot|\mathcal{O}_\ell|)$ and \textsc{Fertility} at
amortised $O(1)$.  Summing over arms, the per-round cost is
$O(|\mathcal{L}|\cdot B\cdot m\cdot t)$.  Since $|\mathcal{L}|$ is a small constant, this
is $O(B\cdot m\cdot t)$, the same class as \textsc{BaseScore} alone.

\paragraph{Router Update.}
Algorithm~\ref{alg:update-app} loops over the $m$ generated functions.
Each iteration performs an $O(1)$ buffer append and one call to
\textsc{Attribute}, whose cost is $O\!\left((P')^{D_{\max}}\right)=O(1)$
for constant $D_{\max}$ and $P'$.  The total per-round cost is
therefore $O(m)$, already subsumed by the \textsc{Attribute} term.

\paragraph{Overall.}
Summing over $T$ rounds, the total router overhead is
\[
  \underbrace{O(B\cdot m\cdot T^{2})}_{\textsc{BaseScore}+\textsc{SelectArm}}
  \;+\;
  \underbrace{O\!\left(m\cdot T\cdot(P')^{D_{\max}}\right)}_{\textsc{Attribute}+\textsc{Update}}
  \;+\;
  \underbrace{O(|\mathcal{L}|\cdot W\cdot T)}_{\textsc{Fertility}}
  \;=\;
  O(B\cdot m\cdot T^{2}),
\]

since $D_{\max}$, $P'$, $W$, and $|\mathcal{L}|$ are all small constants.
The lineage components add only lower-order $O(m\cdot T)$ and $O(|\mathcal{L}|\cdot T)$ terms and do not change the asymptotic complexity relative to the base score computation.

In practice, this is negligible relative to the cost of \llm inference and sandboxed evaluation, which dominate each round by several orders of magnitude.

\subsection{A Note on Regret Guarantees}
\label{app: theoretical_guarantee}
Regret guarantees in the bandit literature rest on two assumptions: that rewards are stationary within an arm, and that arms are independent. Both fail here by construction. The genealogy couples the arms, since one model's output seeds the context another is prompted with (P2), and every arm's score distribution drifts as the population conditioning it improves (P3).
 
A bound can be recovered by positing a parametric form for an LLM's score distribution and for how that distribution shifts across iterations. We, however, deliberately avoid this since no such characterisation exists to the best of our knowledge.